\documentclass[journal]{IEEEtran} 
\IEEEoverridecommandlockouts 

\usepackage[utf8]{inputenc}
\usepackage{graphicx}
\usepackage{authblk}
\usepackage{hyperref} 
\usepackage{caption}
\usepackage{color}
\usepackage{algorithmicx}
\usepackage[ruled,noend,linesnumbered]{algorithm2e}
\SetKwRepeat{Do}{do}{while}
\usepackage[noend]{algpseudocode}
\usepackage{multirow}
\usepackage{graphicx}
\usepackage{subfigure} 
\usepackage{framed}
\usepackage{amssymb}
\usepackage{amsmath }
\usepackage{tcolorbox}
\usepackage{balance}
\usepackage{subfigure}
\usepackage{tikz}
\usepackage[absolute,overlay]{textpos}% by XZ: for the text in the top margin..

\usepackage{fancyhdr}

\newtheorem{example}{Example}[section]

\newcommand{\commentout}[1]{}

\newcommand{\dnnfunction}{\phi}
\newcommand{\rnnfunction}{\psi}

\newcommand{\testsuite}{\mathcal{T}}
\newcommand{\distance}[2]{ ||#1||_{#2}}

\newcommand{\znorm}{{\mathcal N}_z}
\newcommand{\minmaxnorm}{{\mathcal N}_{m}}

\newcommand{\tool}[1]{\textsc{#1}\xspace}
\newcommand{\testRNN}{\tool{testRNN}}

\usepackage{marginnote}
\usepackage{color}

\usepackage{soul}

\DeclareMathOperator*{\argmin}{\arg\!\min}

\title{Coverage Guided Testing for Recurrent Neural Networks}

\author{Wei Huang$^{1}$,
Youcheng Sun$^{2}$, 
Xingyu Zhao$^{1}$
James Sharp$^{3}$, Wenjie Ruan$^{4}$, Jie Meng$^{5}$, and Xiaowei Huang$^{1}$
\thanks{$^{1}$Wei Huang, Xingyu Zhao and Xiaowei Huang are with University of Liverpool, UK.}%
\thanks{$^{2}$Youcheng Sun is with Queen's University Belfast, UK}%
\thanks{$^{3}$James Sharp is with Defence Science and Technology Laboratory, UK}%
\thanks{$^{4}$Wenjie Ruan is with University of Exeter, UK}%
\thanks{$^{5}$Jie Meng is with Loughborough University, UK}%
}

\begin{document}
\begin{textblock*}{20cm}(1cm,1cm)
\centering
\textcolor{red}{Preprint accepted by IEEE Transactions on Reliability. Published version: DOI (identifier) 10.1109/TR.2021.3080664}
\end{textblock*}

\twocolumn

\maketitle
\thispagestyle{empty}
\pagestyle{empty}

\begin{abstract}
Recurrent neural networks (RNNs) have been applied to a broad range of applications, including natural language processing, drug discovery, and video recognition. Their vulnerability to input perturbation is also known. Aligning with a view from software defect detection, this paper aims to develop a coverage guided testing approach to systematically exploit the internal behaviour of RNNs, with the expectation that such testing can detect defects with high possibility. Technically, the long short term memory network (LSTM), a major class of RNNs, is thoroughly studied. A family of three test metrics are designed to quantify 
not only the values but also the temporal relations (including both step-wise and bounded-length) exhibited when LSTM processing inputs. A genetic algorithm is applied to efficiently generate test cases. The test metrics and test case generation algorithm are implemented into a tool \testRNN, which is then evaluated on a set of LSTM benchmarks. Experiments confirm that \testRNN has advantages over the state-of-art tool DeepStellar and attack-based defect detection methods, owing to its working with finer temporal semantics and the consideration of the naturalness of input perturbation. Furthermore, \testRNN\ enables meaningful information to be collected and exhibited for users to understand the testing results, which is an important step towards interpretable neural network testing.
\end{abstract}

\begin{IEEEkeywords}
RNNs, coverage guided testing, coverage metrics, test case generation.
\end{IEEEkeywords}

\section{Introduction}

Feedforward neural networks (FNNs), notably convolutional neural networks (CNNs), are vulnerable in various safety and security scenarios, subject to adversarial attack \cite{szegedy2014intriguing}, backdoor attack \cite{gu2019badnets}, data poisoning attack \cite{10.5555/3327345.3327509}, privacy issues \cite{7958568}, etc. These defects are extensible to recurrent neural networks (RNNs). In this paper, we study the RNN defects, focusing on adversarial samples \cite{alzantot-etal-2018-generating} and backdoor samples \cite{8836465}. These defects will lead a well-trained RNN to mis-predictions. Different from CNNs, RNNs exhibit particular challenges, due to their more complex internal structures and their processing of sequential inputs with a temporal semantics, supported by their internal memory components. 
A generic RNN layer takes a sequential sample $x$ as input, updates its internal state $c$, and generates an output $h$. Other structural components may be required for specific RNNs.   
Given an input $\{x_t\}_{t=1}^n$, the RNN layer will be unfolded with respect to the size $n$ of the input, and therefore each structural component has a corresponding sequence of representations, for example $\{h_t\}_{t=1}^n$. Such a sequence of representations form a temporal evolution. 

Coverage-guided testing has achieved a great success in software defect detection, and has been extended to work with FNNs in e.g., \cite{PCYJ2017,wicker2018feature,ma2018deepgauge,sun2018testing,sun2018concolic}, where a collection of FNN coverage metrics can be found. The definitions of these metrics are based on the structural information, such as the neurons' activations \cite{PCYJ2017,ma2018deepgauge}, the relation between neurons in  neighboring layers \cite{sun2018testing,sun2018concolic}, etc. While existing coverage metrics for FNNs may be adapted to work with RNNs, they are insufficient because they do not work with the internal structures of RNNs and, more importantly, the most essential ingredient of RNNs -- the temporal relation --  is not considered. Moreover, we note that, a few coverage metrics are proposed in \cite{du2019deepstellar} for RNNs, by making simple extensions to those of FNNs without considering the temporal relation and the internal structures (e.g., the important components of RNNs such as gates). \emph{This paper is to develop dedicated coverage metrics for RNNs, to take into account the additional structures and temporal semantics.} 

As suggested in \cite{8805667,dong2019limited}, a test metric does not have to be strongly related to adversarial samples, a specific type of defects corresponding to the robustness requirement of a neural network. This is not surprising, and actually not new (for software testing). 
As stated in \cite{MStestingbook}, a (software) program with high test coverage has more of its source code executed during testing, which suggests it has a lower chance of containing undetected software defects compared to a program with low test coverage. We concur with this view, and suggest that, \emph{instead of identifying a particular type of defects such as adversarial samples, coverage-guided testing is to generate a set of test cases as diversified as possible while preserving the naturalness, in order to exploit the internal behaviour of the neural networks that has real operational impact}. The proposed coverage metrics in this paper are of such desirable features of being \textit{diverse and natural} -- with increased coverage, our approach is more likely to find different types of faulty behaviours (e.g., adversarial samples and backdoor samples) that manifest at multiple small regions in the input space (rather than adversarial samples clustered in one region as what normally attack-based methods find). Especially when the operational profile is unknown or changing, such diversified test cases are of particular importance for improving the delivered reliability \cite{bishop2017deriving} (indeed, spending all the budget on testing one input region that potentially has limited chance to be operated in practice is unwise). Meanwhile, our diversified test cases are ``closer'' to their seeds (points on the RNN's data manifold), compared to other state-of-the-art tools, implying higher chance to be seen in the real-life operation, thus preserving the naturalness.

\paragraph{Contributions}

We first discuss in Section~\ref{sec:problem} why the coverage-guided testing is useful in analysing RNNs and how to reasonably define the effectiveness of a testing framework. We focus on long  short-term  memory  networks (LSTMs), which is the most important class of RNNs, and design three LSTM structural coverage metrics, namely boundary  coverage (BC), step-wise coverage (SC) and temporal  coverage (TC). Simply speaking, TC quantifies the multi-step temporal relation, which describes the internal behaviour on how LSTM cell processing inputs, while BC and SC quantify the value and single-step change of the temporal relation, respectively. We also discussed in Section~\ref{sec:relation} how to position the new metrics against a few closely related techniques such as complete verification techniques, existing metrics, etc. 

We implement the proposed coverage metrics into a prototype 
tool \testRNN\footnote{https://github.com/TrustAI/testRNN}, which includes two algorithms -- a random mutation and a genetic algorithm based targeted mutation -- for test case generation. In particular, targeted mutation uses the coverage knowledge to guide the test case generation. Initially, a random mutation is taken to generate test cases. Once the un-targeted randomisation has been hard to improve the coverage rate, a targeted mutation by considering the distance to the satisfaction of un-fulfilled test conditions is taken to generate corner test cases. 

We conduct an extensive set of experiments over a wide range of LSTM benchmarks to confirm the utility of \testRNN\ and the proposed coverage-guided RNN testing approach from the following aspects:
\begin{enumerate}
    \item diversity of generated test cases (Section~\ref{sec:diversityexp}), with the observations that the LSTM model's functional coverage can be approximated using our structural coverage metrics (Section \ref{sec:function-coverage}) and our metrics complement existing metrics in guiding the exploitation of the input space (Section \ref{sec:existing-coverage}). 
    \item detecting defects (Section~\ref{sec:detectionexp}), with the observations that \testRNN\ can not only find adversarial behaviours for the robustness of RNNs (Section \ref{sec:adversarial-examples}) but also identify backdoor inputs for the security of RNNs (Section \ref{sec:backdoor}). 
    \item usefulness of test case generation (Section~\ref{sec:test_case_generationexp}), with the observation that \testRNN\ is efficient and effective in achieving high coverage rates (Section \ref{sec:test_case_generationexp}).
    \item comparison with dedicated defect detection (Section~\ref{sec:comparison_attack}), with the observation that our test method can find a set of more diversified adversarial samples, and these samples are more likely to occur in real world. 
    \item comparison with state-of-the-art tool DeepStellar (Section~\ref{sec:compare-with-related-work}), with the observations that our metrics are better at guiding the exploitation of the input space and  \testRNN\ may achieve good coverage on the metrics in DeepStellar but not vice versa. 
    \item exhibition of LSTM internal working mechanism (Section~\ref{sec:interpretable}), with the conclusion that semantic meanings behind the test metrics can help users understand the learning mechanism of LSTM model, making a step towards interpretable LSTM testing. 
\end{enumerate}

The organisation of the paper is  as follows. Section~\ref{sec:internalinformation} gives the preliminaries. We will discuss the rationale of coverage-guided testing in Section~\ref{sec:problem}. After this, we present our proposed test metrics in Section~\ref{sec:metrics}. This is followed by discussing  in Section~\ref{sec:relation} how these new metrics are related to the complete verification techniques, existing coverage metrics, and  adversarial defence techniques. We present our test case generation algorithm in Section~\ref{sec:algorithms} and the experimental evaluation in Section~\ref{sec:experiments}. Finally, we review related works in Section~\ref{sec:relatedworks} and conclude the paper in Section~\ref{sec:concls}. 

\section{RNN Preliminaries}\label{sec:internalinformation}

Feedforward neural networks (FNNs) model a function $\dnnfunction:X\rightarrow Y$ that maps from input domain $X$ to output domain $Y$: given an input $x\in X$, it outputs the prediction $y\in Y$. For a sequence of inputs $x_1,\dots,x_n$, an FNN $\dnnfunction$ considers each input individually, that is, $\dnnfunction(x_i)$ is independent from $\dnnfunction(x_{i+1})$. 

By contrast, a recurrent neural network (RNN) processes an input sequence by iteratively taking inputs one by one. A recurrent layer can be modeled as a function $\rnnfunction:X'\times C\times Y' \rightarrow C\times Y'$ such that $\rnnfunction(x_t,c_{t-1},h_{t-1})=(c_t,h_t)$ for $t=1...n$, where $t$ denotes the $t$-th time step, $c_t$ is the cell state used to represent the intermediate memory and $h_{t}$ is the output of the $t$-th time step.  
More specifically, the recurrent layer takes three inputs: $x_t$ at the current time step, the prior memory state $c_{t-1}$ and the prior cell output $h_{t-1}$; consequently, it updates the current cell state $c_t$ and outputs hidden state $h_{t}$.   

RNNs differ from each other given their respective definitions, i.e., internal structures, of recurrent layer function $\rnnfunction$, of which long short-term memory (LSTM) in Equation (\ref{eq:lstm}) is the most popular and commonly used one. 
\begin{equation}
\label{eq:lstm}
\begin{array}{lcl}
f_t & = & \sigma(W_f\cdot [h_{t-1},x_t] + b_f) \\ 
i_t & = & \sigma(W_i\cdot [h_{t-1},x_t] + b_i) \\ 
c_t & = & f_t*c_{t-1} + i_t * \tanh(W_c\cdot [h_{t-1},x_t] + b_c)\\
o_t & = & \sigma(W_o\cdot [h_{t-1},x_t] + b_o) \\ 
h_t & = &  o_t * \tanh(c_t)
\end{array}
\end{equation}
In LSTM, $\sigma$ is the sigmoid function and $\tanh$ is the hyperbolic tangent function; $W$ and $b$ represent the weight matrix and bias vector, respectively; $f_t,i_t,o_t$ are internal gate variables of the cell. In general, the recurrent layer (or LSTM layer) is connected to non-recurrent layers such as fully connected layers so that the cell output propagates further. We denote the remaining layers with a function $\dnnfunction_2:Y'\rightarrow Y$. Meanwhile, there can be feedforward layers connecting to the RNN layer, and we let it be another function $\dnnfunction_1:X\rightarrow X'$. As a result, the RNN model that accepts a sequence of inputs $x_1,\dots,x_n$ can be modeled as a function $\varphi$ such that 
%\begin{equation}
$\varphi(x_1...x_n) = \dnnfunction_2\cdot\rnnfunction(\prod_{i=1}^n \dnnfunction_1(x_i))$.

\iffalse
\begin{figure}
    \centering
    \includegraphics[width=0.85\linewidth]{images/LSTM_cell.pdf}
    \caption{LSTM Cell}
    \label{fig:Cell}
\end{figure}
\fi

\paragraph{Cell structure}\label{sec:internalinformation_cell}
The processing of a sequential input $x = \{x_t\}_{t=1}^n$ with an LSTM layer function $\rnnfunction$, i.e., $\rnnfunction(x)$, can be characterised by gate activations $f=\{f_t\}_{t=0}^n$, $i=\{i_t\}_{t=0}^n$, $o=\{o_t\}_{t=0}^n$, cell states  $c=\{c_t\}_{t=0}^n$, and outputs $h=\{h_t\}_{t=0}^n$. We let ${\mathcal S} =  \{ f, i, o, c, h \}$ be a set of structural components of LSTM, and use variable $s$ to range over ${\mathcal S}$. 

\paragraph{Sequential structure}
Each $s$ represents one aspect of the concrete status of an LSTM cell.  To capture the interactions between multiple LSTM steps, temporal semantics  are often used to understand how LSTM performs \cite{ming2017understanding}. Test metrics in this paper will rely on the structural information such as \emph{aggregate knowledge} $\xi_{t}^{h}$ and \emph{remember rate} $\xi_t^{f,avg}$, as explained below, and their temporal relations.

Output $h$ is seen as short-term memory (as opposed to $c$ for long-term memory) of LSTM. It is often used to understand how information is updated, either positive or negative according to the value of $h_t$. Thus, we have 
\begin{equation}\label{equ:+}
\begin{array}{lcl}
\xi_{t}^{h,+} & = &  \displaystyle \sum \{ h_t(j) ~|~ j \in \{1,\ldots,|h_t|\}, h_t(j) > 0 \}\\
\xi_{t}^{h,-} & = & \displaystyle \sum \{ h_t(j) ~|~ j \in \{1,\ldots,|h_t|\}, h_t(j) < 0 \}\\
\xi_{t}^{h} & = & \displaystyle |\xi_{t}^{h,+} + \xi_{t}^{h,-}|
\end{array}
\end{equation}
Intuitively, $\xi_t^{h}$ represents the \emph{aggregate} knowledge regarding short-term memory. 

The forget gate $f$ is a key factor for long-term memory in LSTM, as it controls whether the aggregate information can be passed on to the next (unfolded) cell or not. The portion of information passed is then measured by $\xi_t^{f,avg}$ as follows.
\begin{equation}\label{equ:avg}
\xi_t^{f,avg}=
\frac{1}{|f_{t}|}\sum_{j=1}^{|f_{t}|} f_{t}(j)
\end{equation}

\begin{figure}[h!]
    \centering
    \includegraphics[width=\linewidth]{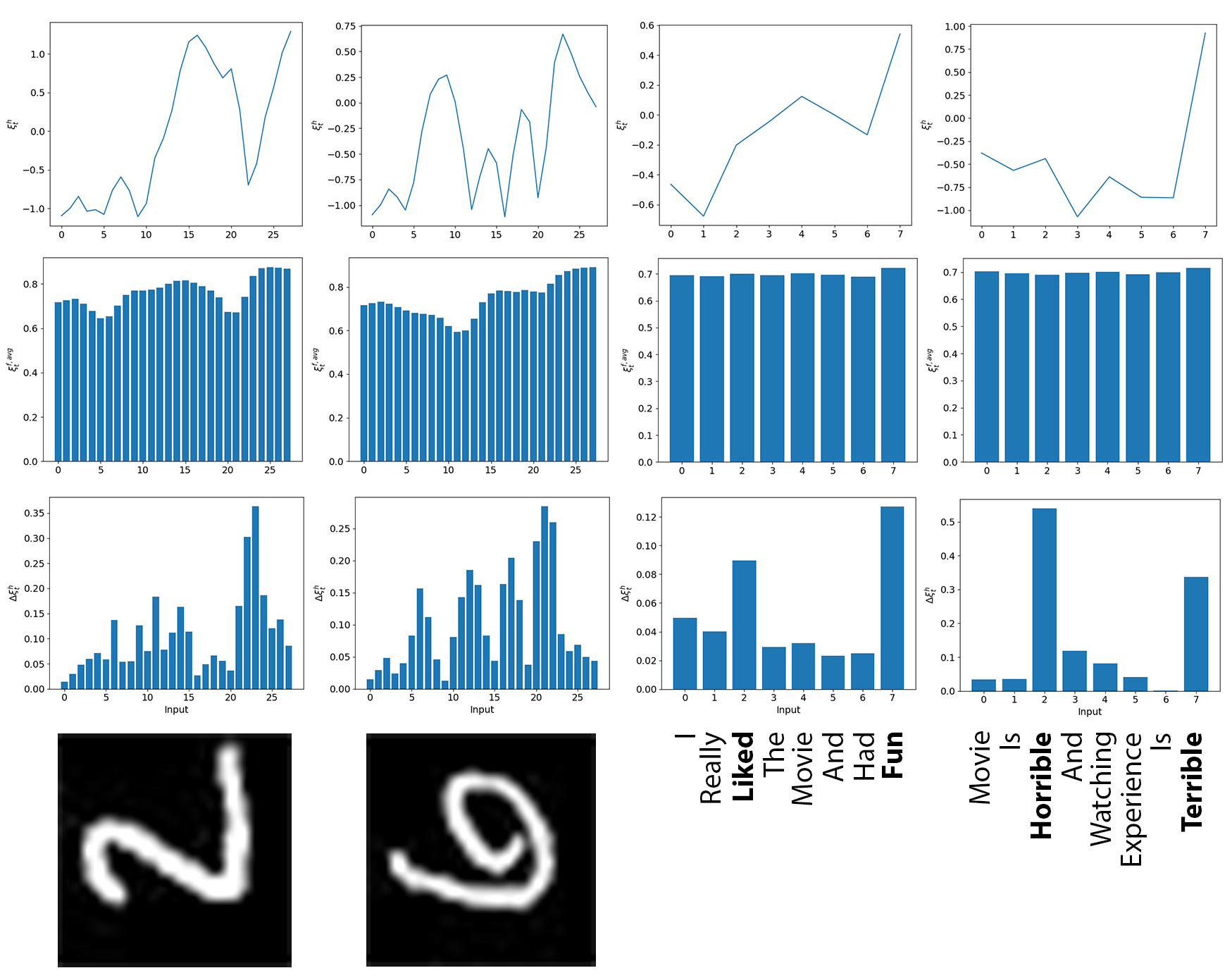}
    \caption{Examples to show how positive and negative elements of output vectors represent the information in MNIST and IMDB models. The x-axis  includes the inputs (bottom row) and the y-axis includes $\znorm(\xi_t^{h})$ (top row),  $\minmaxnorm(\xi_t^{f, avg})$ (second row) and $\minmaxnorm(\Delta\xi_t^{h})$ (third row) values. In MNIST, each column of pixels corresponds to a step in LSTM and in the IMDB model each step represents a word in the movie review.}
    \label{fig:hidden_infor}
\end{figure}

\begin{example}
\label{example-structure}
Fig. \ref{fig:hidden_infor} presents a set of  visualisations to the temporal update of the abstract information. In particular, the top row contains curves for $\znorm(\xi_t^{h})$ and the second row contains curves for $\minmaxnorm(\xi_t^{f, avg})$, changed with respect to the time. The third row visualises the evolution of step-wise change information $\minmaxnorm(\Delta\xi_t^{h})$. $\znorm$ and $\minmaxnorm$ are two normalization function which will be introduced later. 
\end{example}

Let ${\mathcal A}=\{+,-,avg\}$ be a set of symbols representing the abstraction functions as in Eq. (\ref{equ:+}-\ref{equ:avg}). The above can be generalised to work with any $s\in {\mathcal S}$ and $a\in {\mathcal A}$.  For $\xi_{t}^{s,a}$, once given a fixed input $x$, we may write $\xi_{t,x}^{s,a}$.

\section{Problem Statement}\label{sec:problem}

\begin{figure*}
\begin{minipage}{0.59\textwidth}
    \centering
    \includegraphics[width=\textwidth]{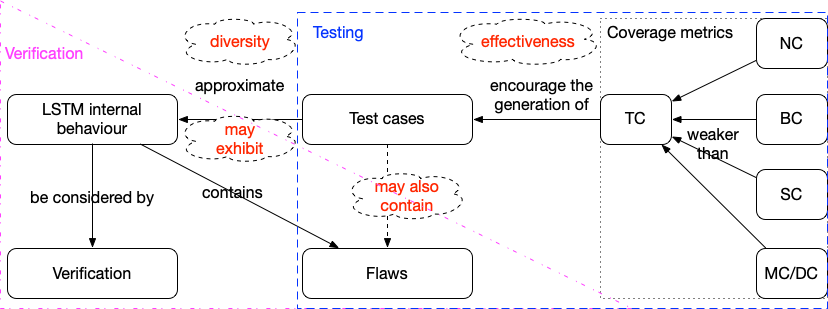}
    \label{fig:framework}
\end{minipage}
\begin{minipage}{0.4\textwidth}
    \centering
    \includegraphics[width=0.8\textwidth]{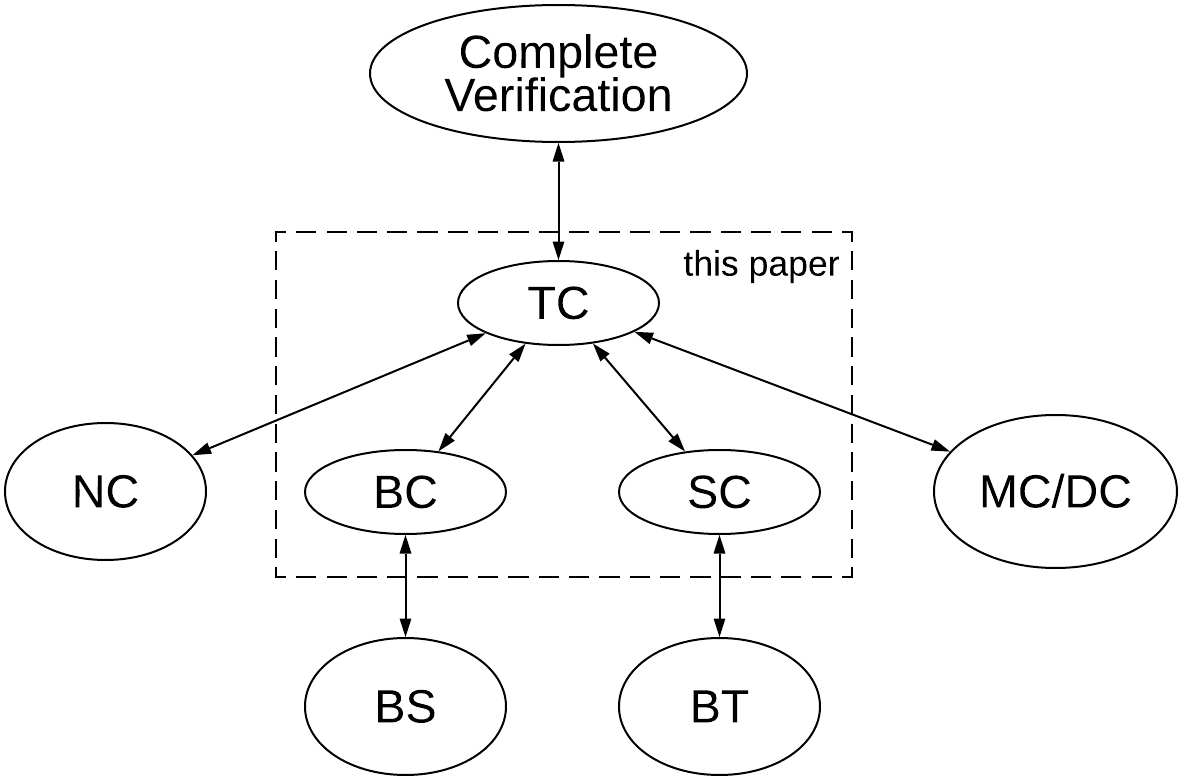}
    \label{fig:relation}
\end{minipage}
    \caption{(Left) Connection of Verification and Coverage Guided Testing Frameworks. Verification and testing overlap on ``Flaws'', representing that they have the same objective. From verification to testing, an approximation is made, i.e., test cases approximate the LSTM internal behaviour. Guidelines (colored with red) are needed to ensure the approximation quality. (Right) Relation between Coverage Metrics. NC: neuron coverage \cite{PCYJ2017}, BS: basic state coverage \cite{du2019deepstellar}, BT: basic transition coverage \cite{du2019deepstellar}, MC/DC: modified condition/decision coverage \cite{sun2018testing}. Arrows represent the ``weaker than'' relation between metrics.}
    \label{fig:theory}
\end{figure*}

This section explains why the coverage guided testing is useful in analysing RNNs and
how to reasonably justify its effectiveness.  
Fig.~\ref{fig:theory}(Left) presents the  connection between verification and testing in this context. While incomplete, testing has been shown practical -- and in many cases sufficiently effective -- in providing assurance to the quality of software. As in Fig.~\ref{fig:theory}, these two approaches overlap on the ``Flaws''. Verification can detect defects because it exhaustively exploits all internal behaviour of RNNs, which include defective behaviours. On the other hand, testing approach uses test cases to approximate -- or sample -- the internal behaviour. Due to the finite sampling, defects may or may not be detected. Therefore, testing needs to be systematic to be effective in defect detection, and coverage-guided testing is one of the main approaches. 

Due to the size and the temporal semantics of RNNs, it becomes important to find a (meaningful) set of metrics to guide the sampling or the test case generation. Our proposed set of coverage metrics plays the role of such guidance -- as suggested in Fig.~\ref{fig:theory}(Left), coverage-guided testing generates a set of test cases to exploit the internal behaviour of the neural networks. We note, such coverage metric does not have to be strongly correlated to adversarial samples \cite{8805667,dong2019limited} (a specific type of defects corresponding to the robustness requirement of a neural network). What really matters is, 
within the testing budget, to find defects that are as diverse and natural as possible so that fixing them would gain maximised impact on the delivered reliability. 

There are two main goals of testing \cite{707695}: debug testing, which probes the software for defects, and operational testing, which is to gain confidence that the software is reliable \cite{zhao_assessing_2020}. The former seeks test cases to excite as many failures as possible (then we may fix the defects behind them), but the test cases normally are not representative of the software’s day-to-day operation. Confidence in the delivered reliability can only be gained by the later, i.e. testing that represents the typical usage (the operational profile) \cite{hamlet_partition_1990}. Coverage-guided testing belongs to the former, while our method should also be designed in the best interest for the operational reliability. That is, our test cases should be more likely generated from the high probability density area on the operational profile, compared to attack-based and other state-of-the-art debug testing methods, so that the defects found are more ``practical'' in the sense that fixing them would effectively improve the operational reliability.

The question is on how effective a specific testing approach is when exploiting the internal behaviour. Below, we provide a few guidelines by evaluating the connections of entities in Fig.~\ref{fig:theory}(Left) (shown with red color).

First, the test cases are required to be \emph{diversified and natural}, so as to cover the LSTM internal behaviour comprehensively. This is to avoid the test cases being clotted together in a small region of the input space (representing a certain type of defects) and lose the ability of finding other defects manifested in different regions. However, it is easy to generate diversified test cases by ``forcing diversity'' (e.g., by selecting inputs that maximise the average inter-point distance). Thus diversity criteria is only sensible when paired with naturalness -- tests cases should not be far from the RNN's data manifold (i.e., potentially high density area of its future optional profile). Second, the test cases can reveal defects. While it is hard to establish strong correlation between test metrics and a specific type of defects, it is still a reasonable request that the generated test cases \emph{reveal} defects as many as possible. Third, the test case generation algorithm needs to be effective, in terms of its ability in improving the coverage rate with diversified and natural test cases. Finally, to show that the generated test cases are sufficiently representative, we may use the test cases to exhibit the working mechanism of LSTM. 

Moreover, given the complexity of the internal behaviour in RNNs, we believe a set of coverage metrics are needed to ensure that the above guidelines can be achieved. In an ideal case where the testing budget is sufficient, our metrics in the paper may complement -- instead of replace -- others, and vice versa. These guidelines will be used when designing our experiments in Section \ref{sec:experiments}.

\section{LSTM Test Coverage Metrics}\label{sec:metrics}

In this section, we present a family of three coverage metrics (BC, SC and TC) for the testing of LSTM models. These metrics take into account both the values of structural information $\xi_{t}^{s,a}$ for $s\in {\mathcal S}$ and $a\in {\mathcal A}$ as in Eq. (\ref{equ:+}-\ref{equ:avg}) and their step-wise and bounded-length temporal relations. We utilize two normalization methods for the convenience of determining thresholds, independent of specific dataset. $\znorm$ and $\minmaxnorm$ are z-score and min-max normalization function defined as below:
$$
\znorm(\xi) = \frac{\xi-\mu}{\sigma},\quad \minmaxnorm(\xi) = \frac{\xi - min}{max-min}
$$
where parameters $\mu$, $\sigma$, $min$ and $max$ can be derived from the training dataset. The z-score normalization is suitable for preprocessing test conditions quantifying the relations between different features, e.g., TC, while min-max normalization is better for the test conditions to hit the large activation values, e.g., BC and SC. Given a time series of length $n$, we can choose the sequence of interest $[t_1, t_2]$ ($t_1 \geq 1$ and $t_2 \leq n$) to implement the following coverage metrics. 

\paragraph{Boundary Coverage (BC)}\label{sec:gatecoverage}
Boundary values are often regarded as important cases in software testing, as they could exploit extreme software behaviours. 
We therefore define BC for depicting test conditions that cover the boundary values of the LSTM data flow as follows.
$$\{\minmaxnorm(\xi_{t}^{s,a}) \geq \alpha_{max},\,\,\,\minmaxnorm(\xi_{t}^{s,a}) \leq \alpha_{min} ~|~t\in \{t_1...t_2\}\}$$
Thresholds $\alpha_{max}$ and $\alpha_{min}$ are chosen from interval $[0, 1]$. The $min$, $max$ values can be estimated using values computed over the training dataset $\{\xi_{t,x}^{s,a}~|~t \in \{t_1...t_2\}, x \in \mathcal{D}_{train}\}$. 

\begin{example}
Suppose that there is a test condition $\minmaxnorm(\xi_{t}^{i,avg})> 0.9$. It requires that the $\minmaxnorm(\xi_t^{i,avg})$ value is greater than threshold $0.9$. Intuitively, this condition exercises LSTM's learning ability on the input at time $t$. As Eq. \ref{eq:lstm} shows, the input gate $i$ controls how much information from the input is received by the network: $\minmaxnorm(\xi_{t}^{i,avg}) = 1$ implies that all its information is added to the long-term memory $c$ and $\minmaxnorm(\xi_{t}^{i,avg}) = 0$ implies that no input information is added.
\end{example}

\paragraph{Step-wise Coverage (SC)}\label{sec:cellcoverage}
SC characterizes the temporal
changes between connected cells. We use 
$\Delta\xi_t^{s}=|\xi_{t}^{s,+}-\xi_{t-1}^{s,+}|+|\xi_{t}^{s,-}-\xi_{t-1}^{s,-}|$ to outline the maximum change of the structural component $s\in {\mathcal S}$ at time $t$. E.g., $\Delta\xi_t^{h}$ is the change of 
short-term memory at time $t$. Then, the SC test conditions are defined as follows.
$$\{ \minmaxnorm(\Delta\xi_t^{s}) \geq \alpha_{SC}~|~t\in \{t_1...t_2\}\}$$
This set defines test conditions for LSTM's step-wise updates that exceed a threshold $\alpha_{SC}$. Parameters for $\minmaxnorm$ are derived from $\{\Delta\xi_{t,x}^{s}~|~t\in \{t_1...t_2\}, x \in \mathcal{D}_{train}\}$.

\begin{example} 
The intuition behind step-wise coverage is to capture these significant inputs to the LSTM.
As shown by the sentiment analysis LSTM example in Fig. \ref{fig:hidden_infor} (Right), given two inputs, sensitive words ``like", ``horrible", ``fun"  trigger greater $\minmaxnorm(\Delta\xi_t^{h})$ values than  words  ``movie'', ``really'', and ``had''. 
\end{example} 

\paragraph{Temporal Coverage (TC)}\label{sec:sequentialcoverage}
While the power of LSTM comes from its ability to memorize values over arbitrary time intervals, its test metrics need to ensure that the temporal patterns of memory updates are fully tested. This is essentially a time series classification problem and is intractable. In this part, we define test conditions to exploit temporal patterns of bounded length. Different from dynamic systems where the temporal relation can be infinite~\cite{clarkebook}, the temporal relations in RNNs are always finite, because of the finite-sized input. Therefore, the bounded length does not lower the expressiveness of the test conditions. In particular, to facilitate the enumeration of all test conditions, we refer to symbolic aggregate approximation (SAX) \cite{lonardi2002finding}  to convert any complicated time series of length $v$ ($v$ is usually a large number) into a symbolic sequence of length $w$ ($v >> w$).

\begin{figure}
    \centering
    \includegraphics[width=0.95\linewidth]{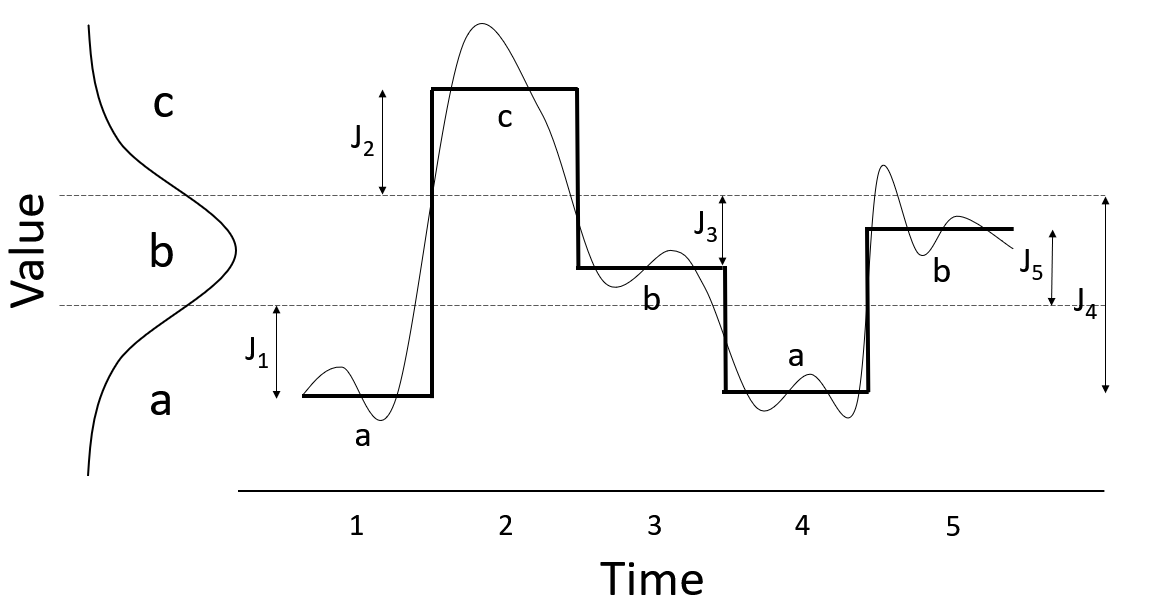}
    \caption{Illustration of projecting a temporal curve (Gaussian distribution) into a sequence of symbols $acbab$.} 
    \label{fig:mnist_sax}
\end{figure}

First of all, given any temporal curve $\xi^{s} = \{\xi_{t}^{s}\}_{t=t_1}^{t_2}$, we can reduce the dimension of temporal sequence from $v = t_2 - t_1$ to $w$ following Piece-wise Aggregate Approximation (PAA). 
 \begin{equation}\label{equ:paa}
 \hat{\xi}_j^{s} = \frac{w}{v} \sum\limits_{t=\frac{v}{w}(j-1)+t_1}^{\frac{v}{w}j+t_1} \xi_t^{s}
\end{equation}
The main idea of PAA is to approximate the original time series by splitting them into $w$ equal sized segments and average the values in each segment. For example, the temporal curve in Fig.~\ref{fig:mnist_sax} is split into $w = 5$ dimensions. The new approximated curve is denoted as $\hat{\xi}^{s} = \{\hat{\xi}_{j}^{s}\}_{j=1}^{w}$.

Then, we can define the symbolic representation for the temporal curve after dimensionality reduction. We start from z-normalizing $\hat{\xi}^{s}$ and discretising $D(\znorm(\hat{\xi}^{s}))$ -- the domain of $\znorm(\hat{\xi}^{s})$ -- into a set $\Gamma$ of sub-ranges. This discretization can refer to the distribution of $\znorm(\hat{\xi}_j^{s})$, which can be estimated by conducting probability distribution fitting over the training dataset (since $\hat{\xi}^{s}$ is z-normalized, $\znorm(\hat{\xi}_j^{s})$ is subject to the standard normal distribution).
 Then, every normalized time series $\{\znorm(\hat{\xi}_j^{s})\}_{j=1}^w$ can be represented as a sequence of symbols in the standard way. For example, in Fig.~\ref{fig:mnist_sax}, the continuous space of $\znorm(\hat{\xi}^{s,a})$ is split into a set of three sub-ranges $\Gamma=\{a,b,c\}$.

\begin{figure*}
    \centering
    \includegraphics[width=\linewidth]{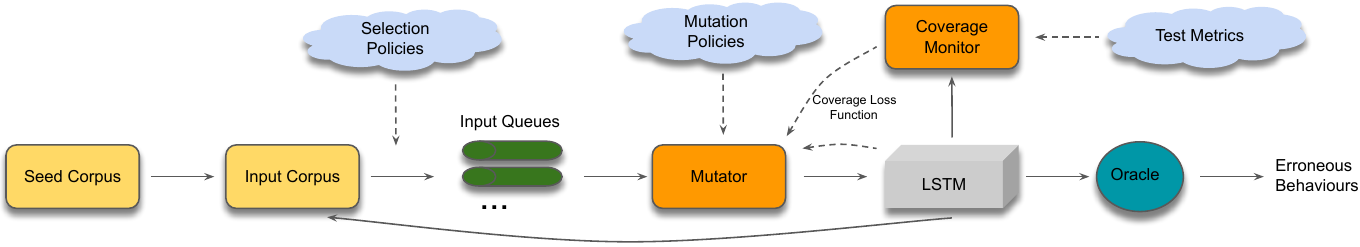}
    \caption{Coverage-guided LSTM testing in \testRNN}
    \label{fig:arch}
\end{figure*}

Finally, test conditions from TC for covering a set of symbolic representations across multiple time steps $[t_1,t_2]\subseteq [1,n]$ can be expressed as follows.
\begin{equation}\label{equ:sequencecoverage}
 \{\ell_{1}\ell_{2}...\ell_{w}~|~ \ell_j \in \Gamma, j \in [1,w]\}
\end{equation}
Essentially, TC requests the testing to meet a set of temporal patterns for a specific time span $[t_1,t_2]$. The total number of temporal patterns for TC to cover is $|\Gamma|^w$. We remark that with the help of SAX, test conditions in TC is scalable in tackling the complexity of time series.

\begin{example} 
Fig.~\ref{fig:hidden_infor} (top row) demonstrates the temporal curve of hidden memory for each input across a selected time span. The curve is for $\znorm(\xi_t^{h})$ and it is a clear illustration on the information processing of LSTM for each input.
Fig.~\ref{fig:mnist_sax} further shows how  a time series is converted into its symbolic representation $acbab$ with $\Gamma = \{a,b,c\}$ and $w=5$. 
\end{example}

\section{Relation with RNN Defects}
\label{sec:relation}

The aforementioned three coverage metrics encourage the exploration of the LSTM internal behaviour, which is helpful in detecting the RNN defects. In this section, we discuss the rationale behind by referring to the general relation (Fig.~\ref{fig:theory}(Right)) between our new coverage metrics, the complete verification and a few existing coverage metrics. All discussions are supported by our experiments in Section~\ref{sec:experiments}.

\paragraph{Comparing with Complete Verification Techniques}\label{sec:comveri} For an input of length $n$, we denote the collection of curves (as in Fig.~\ref{fig:mnist_sax}) for $f, i, o$ as $Curv$. It precisely identifies an output (extracted features of the LSTM layer) and corresponds with a set of inputs (which cannot be differentiated by the LSTM layer). Let $C_f$, $C_o$, and $C_i$ be the (possibly infinite) set of possible curves for their respective gates $f,o,i$. We have $Curv = C_f\times C_o \times C_i$. We also have curves $C_h$ and $C_o$, which can be obtained from $Curv$ according to Equation \eqref{eq:lstm}. 

A complete verification method determines if there is an input that can lead to any unexpected behaviour. That is, it is equivalent to determine if there is a combined curve in $Curv$ that leads to the unexpected classification\footnote{As shown in the experiments, an unexpected classification can be normal mis-classification or caused by e.g. backdoor attacks and adversarial inputs.}. To this end, TC (and its test case generation) can be seen as an approach to exhaustively, but discretely, explore one of the curve sets $C_s$ for $s\in {\mathcal S}$. Therefore, while the combination of TC working on different gates may provide a complete verification, in general the exploration of internal behaviour through TC is a necessary, but insufficient, approach for verification. 

\paragraph{Comparing with Other Coverage Metrics}
\label{sec:comparison_metrics}
The purpose of TC is to encourage the exploration of either $C_f$, $C_o$, or $C_i$ with the generated test cases. However, such exploration may be computational intensive -- the complexity is exponential with respect to the length $n$. Since BC and SC do not consider the temporal relation between steps or boundary values, they are computationally more manageable. Assuming that under ideal parameter settings (e.g., thresholds $\alpha_{max}$ and $\alpha_{SC}$ and the set $\Gamma$ of symbols), we say that a metric $A$ is weaker than another $B$ if for any test suite, it cannot have a lower coverage rate w.r.t. $A$ than that of $B$. It is not hard to see that, both BC and SC are weaker than TC, and BC and SC are incomparable, as shown in Fig.~\ref{fig:theory}(Right). 

Besides the new BC and SC, TC is stronger than existing coverage metrics that are originally proposed for CNNs. For example, neuron coverage (NC) \cite{PCYJ2017}, which requires the coverage of neurons whose value is over a threshold (e.g., 0 for ReLU activation function), can be adapted to work with say the gate value or hidden state value. In this case, it is weaker than TC and incomparable with SC. Although NC and BC have similar formal expressions, BC concerns the boundary value rather than a value that indicates the activation status. For the MC/DC metrics \cite{sun2018testing}, they can be adapted to work between time steps in the new context of RNNs. With such adaptation, MC/DC encourages the exploration of relations between time steps, and therefore are weaker than TC. 

DeepStellar \cite{du2019deepstellar} abstracts the evolution of hidden states of an RNN into a discrete-time Markov chain (DTMC) before considering state and transition coverage. Its basic state coverage (BS) and basic transition coverage (BT) are designed to cover the possible state values and possible transitions. Given that the DTMC is an abstraction of the curves $C_h$, BS and BT are both weaker than BC and SC, respectively.

Remarkably, the relations in Fig.~\ref{fig:theory}(Right) are based on theoretical analysis under ``ideal parameter settings''. They do not hold for any parameter settings. In our experiments in Section~\ref{sec:experiments}, we might observe the coverage rates on the same test set and the aforementioned relations are not in alignment, because of the specific parameter settings used (cf. Table~\ref{tab:parameters}).

\paragraph{Defence Techniques for Robustness and Security} Some effective adversarial defence techniques, e.g. \cite{crecchi2019detecting}, are based on the observation that the adversarial samples exhibit different internal behaviour to those behaviour of training data samples. For security concerns like backdoor attack, activation patterns are also considered in the detection techniques such as \cite{chen2018detecting}. Consequently, with the exploration of more RNN internal behaviour other than those appeared in the training data, it is more likely that RNNs defects will be exposed. 

\section{Coverage Guided Test Case Generation}\label{sec:algorithms}
Coverage metrics in Section \ref{sec:metrics} define test conditions that request particular patterns of long/short-term 
updates of abstracted information across multiple LSTM time steps. Given an LSTM network and a specific test metric, the \textbf{coverage rate} denotes the percentage of test conditions that have been satisfied over a set of test cases, i.e., test suite. To more efficiently achieve high coverage rate, in this section, we develop the coverage guided test case generation, as outlined in Fig. \ref{fig:arch}. We remark that, although focus of this paper is LSTM, the proposed testing approach (including both test metrics and tests generation) can be extended to work with other kinds of RNNs which use customized recurrent layer structures. 

The \testRNN test case generation algorithm is detailed in Alg. \ref{alg:xmainAlgorithm}. The test suite $\testsuite$ is initialized with $\testsuite_0$, a corpus of seed inputs (Line $1$). New test cases are generated by mutating seed inputs. It keeps the traceability of test cases via a mapping $orig$ that maps each test case generated back to its seed origin. 

The main body of Alg. \ref{alg:xmainAlgorithm} is a loop (Lines $5$$-$$10$) that iterates unless some target coverage level is reached (Line 4). At each iteration, a test input $x$ is selected from the input corpus $\mathcal{T}$ (Line $5$), and it is mutated following the  pre-defined mutation function $m$ (Line 6). Newly generated test inputs are added into $\mathcal{T}$ (Line 7), where they are %further selected and 
queued for the next iteration. If the generated test case does not pass the oracle (Section~\ref{sec:oracle}), it represents a defect and it is added to $\testsuite_{adv}$ (Lines 9-10).  

\begin{algorithm}[tbhp]
\caption{\testRNN Algorithm}
\label{alg:xmainAlgorithm}
\KwIn{\\
$\dnnfunction$: RNN to be tested \\
$\testsuite_0$: a set of seed inputs \\
$m$: a mutation function \\
$r_{oracle}$: oracle radius
}

\KwOut{\\
$\testsuite$: a set of test cases \\
$\testsuite_{adv}$: a set of discovered adversarial samples

}
$\testsuite \leftarrow \testsuite_0$ \\
$orig\leftarrow dict()$\\
$orig[x] \leftarrow x$ for all $x\in \testsuite_0$ \\ 
\While{coverage rate is not satisfied}
{
$x\leftarrow$ select an element from $\testsuite$\\
$x' \leftarrow m(x)$\\
%${\cal U}={\cal U}\cup \{x\}$
$\testsuite\leftarrow \testsuite \cup \{x'\}$ \\
$orig[x']\leftarrow orig[x]$ \\

\If{$\distance{orig(x)-x'}{2} \leq r_{oracle}$ and $\dnnfunction(x) \neq \dnnfunction(x')$
}
{
    %$x' = m(x)$\\
    $\testsuite_{adv} \leftarrow \testsuite_{adv} \cup \{x'\}$\\
}
}
return $\testsuite, \testsuite_{adv}$
\end{algorithm}

\subsection{Selection Policies and Queuing}

Not all inputs in the corpus $\mathcal{T}$ are equivalently important, and they are ranked once added to the input queue (as illustrated in Fig. \ref{fig:arch}). When sorting queuing inputs on $\mathcal{T}$ for the Mutator engine, $\testRNN$ particularly prioritizes two kinds of test inputs: those that are promising in leading to the satisfaction of un-fulfilled test conditions and those that can trigger erroneous behaviours. 

Thanks to its modular design, new selection policies can be easily integrated into \testRNN as plug-ins (as indicated by cloud shapes in Fig. \ref{fig:arch}). The design of \testRNN also features  its high parallelism. The use of dynamically allocated input queues further optimises its runtime performance.

\subsection{Mutation Policies}\label{sec:mutation}
The Mutator engine lays at the core  of \testRNN. In particular, there are two types of mutation function $m$ in Alg.~\ref{alg:xmainAlgorithm}: random mutation $m_{rnd}$ and targeted mutation $m_{targ}$. 

\paragraph{Random Mutation}
When the LSTM input has continuous values (e.g., image input), Gaussian noises with fixed mean and variance are added to the input. Meanwhile, for discrete value input (e.g., IMDB movie reviews for sentiment analysis), a set of problem-specific mutation functions ${\mathcal M}$ are defined. The detail is in the experiment set-up (Section \ref{sec:models}).

\paragraph{Targeted Mutation}
The targeted mutation is based on genetic algorithm for test case generation.
Genetic algorithm is an evolutionary approach inspired by the process of natural selection. Mutations are selected only when they improve over the existing test cases on some pre-defined fitness function. The implementation of genetic algorithm comprises of four steps: initialization, selection, crossover and mutation. The last three steps are running iteratively till the solution is found.

\textbf{Initialization}. Firstly, we initialize the population by choosing a test case from the previous running cases. The test case is very close to the satisfaction of test condition.

\textbf{Selection}. Next, we select best few test cases from the population to the mating pool, evaluated by the fitness function. For the three classes of test conditions (BC, SC, TC) with respect to some  $s\in {\mathcal S}$ and $a\in {\mathcal A}$, we define the following fitness function as the distance to their respective targets, e.g., 
$$
\begin{array}{l}
J_{BC}(x) =  \alpha_{max} -\minmaxnorm(\xi_{x,t}^{s,a}) \\
J_{SC}(x) =  \alpha_{SC} - \minmaxnorm(\Delta\xi_{x,t}^{s}) \\
J_{TC}(x) =  \sum\limits_{j=1}^{w} dist(\znorm(\hat{\xi}_{x,j}^{s}), u_j) 
\end{array}
$$
where $t, t_1, t_2, j$ are time steps that can be inferred from the context. $u_j = [u_l, u_r]$ is the interval of sub-range, represented by some symbol in $\Gamma$. The fitness of temporal curve to the targeted symbolic curve is to calculate the Manhattan distance, the absolute difference between structure value $\znorm(\hat{\xi}_{x,j}^{s})$ and the symbolic interval $u_j$ is
$$
dist(\znorm(\hat{\xi}_{x,j}^{s}), u_j) = \begin{cases}
            \znorm(\hat{\xi}_{x,j}^{s}) - u_r  & \text{if } \znorm(\hat{\xi}_{x,j}^{s}) > u_r \\
             u_l - \znorm(\hat{\xi}_{x,j}^{s})  & \text{if } \znorm(\hat{\xi}_{x,j}^{s}) < u_l \\
             0  & \text{else } 
       \end{cases} \quad
$$

Intuitively, the fitness function (also called coverage loss) $J(x)$ is estimates the distance to the satisfaction of an un-fulfilled test condition. $J(x) \leq 0$ means that the test condition is covered. By generating test cases with the objective of gradually minimising the loss, the targeted mutation is essentially a greedy search algorithm.

\begin{example}
In Fig. \ref{fig:mnist_sax}, the symbolic representation of temporal curve is $acbab$; the fitness of the curve to test condition $bbcca$ can be calculated as $J_{TC} = J_1 + J_2 + J_3 + J_4 + J_5$.
\end{example}

\textbf{Crossover}. Crossover is trivial in our test case generation for RNNs. This is mainly because the inputs to RNNs are usually discrete, which means the offspring of two parents by crossover (like exchanging chromosome) may be invalid due to the undefined semantic meanings. To avoid the validity issue, we skip this step and using mutation methods directly.

\textbf{Mutation}. We randomly mutate the test cases in the mating pool with user-defined function ${\mathcal M}$ in order to generate a new population for the next iteration. The previous population is replaced with the new one. It should be noticed that the parents in the mating pool are also added to the new population to make sure that  solutions are towards good directions during the iterations.

\begin{algorithm}[tbhp]
\caption{Targeted Mutation}
\label{alg:muta_grad}
\KwIn{\\
$J$: fitness function \\
%$\alpha$: thresholds \\
$\gamma$: maximun iteration number \\
$k$: number of parents for mating pool \\
$n$: number of offsprings mutated from one parent \\
$m_{rnd}$: a random mutation function \\
$x'$: a test case that is the closest to satisfy test condition
}

\KwOut{\\
$x'_{new}$: a test case covering new test condition
}

$itr \leftarrow 0$ \\
%add $x'$ to the population set $P$ \\
$P\leftarrow \{x'\}$\\
\While{test condition is not satisfied in $P$ and $itr < \gamma$ }
{
sort individual $x \in P$ according to fitness $J(x)$ \\
%select the best $k$ individuals in $P$ and add them to $P^*$ \\
$P^*\leftarrow $ highest sorted $\min\{k,|P|\}$ individuals in $P$\\
%$P^* \cup m_{rnd}(P^*,n)$ \\
$P \leftarrow P^* \cup m_{rnd}(P^*,n)$\\
$itr \leftarrow itr + 1$
}
$x'_{new} \leftarrow \argmin_{x \in P} J(x)$ \\ %, for $x \in P$\\
return $x'_{new}$
\end{algorithm}

The targeted mutation $m_{targ}$ is shown in Algorithm \ref{alg:muta_grad}. At each iteration, we choose $k$ (or $|P|$, whichever is smaller) best individuals in the population $P$. % (except for the first iteration where we only have a seed). 
Each individual is utilized to generate  $n$ test cases via the random mutation function $m_{rnd}$. The old population $P$ will be replaced with the mutants along with their $k$ parents in $P^*$. The whole process is repeated until the test condition is met or the maximum iteration is exceeded.

\subsection{Test Set Evaluation}
\paragraph{Test Oracle}\label{sec:oracle}
Test oracle determines if a test case passes or fails. We define a set of norm-balls, each of which is centered around a data sample with known label. The radius $r_{oracle}$ of norm-balls intuitively means that a human cannot differentiate between inputs within a norm ball. In this paper,
Euclidean distance, i.e. $L^2$-norm $\distance{\cdot}{2}$ is used. A test case $x'$ is said to not pass the oracle if (1) $x'$ is within the norm-ball of some known sample $x$, i.e., $\distance{x-x'}{2} \leq  r_{oracle}$, and (2) $x'$ has a different classification from $x$, i.e.,  $\varphi(x)\neq \varphi(x')$. Take the definition in \cite{szegedy2016rethinking}, a test case does not pass the oracle is an adversarial sample. We use \textbf{adversary rate} to denote the percentage of test cases that do not pass the oracle.

\paragraph{Diversity of Test Set}\label{sec:diversity_measure}

More diversified test cases will explore more input space and thus are more likely to uncover different defects. Unfortunately, a unified, accurate way to measure diversity may not exist. We consider the following three intuitive, yet measurable proxies to the diversity. 

First, diversity can refer to the number of categories the generated test cases belonging to. Intuitively, if the labels of two test cases are different, they are dissimilar and more diversified than two test cases with the same label. Second, test metrics (e.g., neuron coverage, SC, BC and TC) are to guide the exploitation of different internal behaviours of RNNs (cf. Section~\ref{sec:metrics}). Therefore, a test set that can achieve higher coverage on more test metrics is more diversified than the other. Third, if the distance in input space, measured with  $L_1$, $L_2$, and $L_\infty$ norm, can represent the semantic similarity between test cases, we may define the diversity by quantifying the relative positions of test cases to the seed input. Suppose a test set $\testsuite$ contains $n$ test cases, generated from a seed $x_0$, the angular-based diversity measure \cite{gong2019diversity} of $\testsuite$ is %
\begin{equation}
    Diversity(\testsuite) = -(\sum_{i, j = 1}^{n} \frac{<x_i-x_0, x_j-x_0>}{||x_i-x_0||~||x_j-x_0||}) /n^2
\end{equation}
This diversity measure is formed by the cosine similarity \cite{yu2011diversity} and bounded by $[-1,1]$. Since the test cases are generated by adding small perturbations to the seed input, the angular-based diversity is to measure if the test cases are uniformly distributed around the seed input $x_0$. A larger $Diversity(\testsuite)$ represents a more diversified $\testsuite$.

\section{Evaluation}\label{sec:experiments}

We evaluate our \testRNN approach with an extensive set of experiments from the following aspects: (1) the diversity of test cases generated under the guidance of the coverage metrics (Section~\ref{sec:diversityexp}), (2) the ability of detecting RNN defects (Section~\ref{sec:detectionexp}), (3) the effectiveness of the test case generation algorithms (Section~\ref{sec:test_case_generationexp}), (4) the advantages over state-of-the-art attack tool \cite{papernot2016crafting} (Section~\ref{sec:comparison_attack}), (5) the difference from state-of-the-art RNN testing tool DeepStellar \cite{du2019deepstellar}  (Section~\ref{sec:compare-with-related-work}), and (6) the exhibition of LSTM internal working mechanism (Section~\ref{sec:interpretable}). Specifically, we study the following research questions (RQs):

\begin{itemize}
  \item \textbf{RQ1}: will the exploitation of internal behaviour lead to the testing of different LSTM functions? 
  \item \textbf{RQ2}: are our new metrics needed when we already have existing metrics? 
  \item \textbf{RQ3}: will the exploitation of internal behaviour lead to the detection of adversarial samples? 
  \item \textbf{RQ4}: will the exploitation of internal behaviour lead to the identification of backdoor attacks? 
  \item \textbf{RQ5}: Can the test case generation algorithm achieve high coverage for the proposed test metrics?
  \item \textbf{RQ6}: What are the advantages of \testRNN over attack-based methods for detecting adversarial samples?
   \item \textbf{RQ7}: What are the similarities and differences between DeepStellar and \testRNN?
    \item \textbf{RQ8}: Are the testing results based on the proposed test metrics helpful on making LSTM interpretable?
\end{itemize}

All the experiments are run on a desktop with Intel(R) Core(TM) i7 CPU @ 3.80\,GHz and 16\,GB Memory.

\subsection{Experimental Setup}\label{sec:models}
\subsubsection{RNNs under Evaluation}
Our experiments are conducted on a diverse set of LSTM benchmarks, including:
\paragraph{MNIST Handwritten Digits Analysis by LSTM}
The MNIST database, containing a set of $60,000$ grey-scale images of size $28$$\times$$28$, is used to train a RNN model with 4 layers. The first two layers are LSTM layers, which are correspondingly connected and fed with rows of input images. That is, each input image is encoded as the row vector of shape $(28,128)$ by the first LSTM layer, and then second layer will do further processing to output an image vector representing the whole image. Finally, two fully-connected layers with ReLU and SoftMax activation functions respectively, are used to process the extracted feature information to get the classification result. The model achieves $99.2\%$ accuracy in training dataset ($50,000$ samples) and $98.7\%$ accuracy in the default MNIST test dataset ($10,000$ samples). 

\paragraph{Sentiment Analysis by LSTM}
 
The sentiment analysis network has three layers, i.e., an embedding layer, an LSTM layer, and a fully-connected layer, with $213301$ trainable parameters. The embedding layer takes as input a vector of length $500$ and outputs a $500$$\times$$32$ matrix, which is then fed into the LSTM layer. Subsequently, there is a fully-connected layer of $100$ neurons.

\paragraph{Lipophilicity Analysis by LSTM} 

We trained an LSTM regression network on a Lipophilicity dataset from the MoleculeNet \cite{WRFGGPLP2017}.
The model has four layers: an embedding layer, an LSTM layer, a dropout layer, and a fully connected layer. The input is a SMILES string representing a molecular structure and the output is its prediction of Lipophilicity. A dictionary is used to map the symbols in the SMILES string to integers. We use the length of the longest SMILES in training dataset as the number of cells for the LSTM layer. Similar to text processing in the IMDB model, short SMILES inputs are padded with 0s to the left side. We use the root mean square error (RMSE) as the measurement of model accuracy. Our trained model achieves RMSE = 0.2371 in training dataset and RMSE = 0.6278 in test dataset, which are better than the traditional and convolutional methods used in \cite{WRFGGPLP2017}.

\paragraph{Video Recognition for Human Behaviour}
A large scale VGG16+LSTM network is trained over the UCF101 dataset \cite{ucf101}. VGG16, a CNN for ImageNet, extracts features from individual frames of a video. Then, the sequence of frame features are analysed by LSTM layer for classification.

\subsubsection{Test Metrics}

We conduct experiments on several concrete test metrics, i.e., BC (for $\xi_t^{f,avg}$), SC (for $\Delta\xi_{t}^{h}$), and TC (for $\xi_{t}^{h}$). The configuration of thresholds are presented in Table \ref{tab:parameters}. Although the proposed three test metrics can be applied to every internal vector of LSTM cell, like $f, i, o, c, h$, the current settings represent better semantic meanings. The interpretation of the testing results is discussed in Section \ref{sec:interpretable}. 

\subsubsection{Input Mutation}
For MNIST model, we add Gaussian noise to input image and round off the decimals around 0 and 1 to make the pixel value stay within the value range. 

The input to IMDB model is a sequence of words, on which a random change may lead to an unrecognisable (and invalid) text paragraph. To avoid this, we take a set ${\mathcal M}$ of mutants from the EDA toolkit \cite{wei2019eda}, which was originally designed to augment the training data for improvement on text classification tasks. This ensures the mutated text paragraphs are always valid. In our experiments, we consider four mutation operations, i.e., ${\mathcal M}$ includes (1) Synonym Replacement, (2) Random Insertion, (3) Random Swap, and (4) Random Deletion. The text meanings are reserved in all mutations. 
For Lipophilicity model, we take a set  ${\mathcal M}$ of mutants  which change the SMILES string without affecting the molecular structure it represents. The enumeration of possible SMILES for a molecule is implemented with the Python cheminformatics package RDkit \cite{rdkit}. Each input SMILES string is converted into its molfile format, based on which the atom order is changed randomly before converting back. There may be several SMILES strings representing the same molecular structure. The enumerated SMILES strings are the test cases.

For UCF101 model, we add Gaussian noise to the original video frames instead of the feature inputs to the LSTM layer.

\subsubsection{Oracle Setting}

We use one fixed oracle radius for each model across all experiments. For continuous inputs, like images and videos, we calculate the euclidean distance as the measurement of perturbation. For the discrete inputs, like text, We refer to the alpha parameter provided by the EDA toolkit, which approximately means the percent of words in the sentence that will be changed. That said, the $r_{oracle}$ for each RNN are listed in Table~\ref{table:model}. Note,  we let $r_{oracle}=\text{None}$ for the Lipophilicity model, suggesting that no constraint is imposed on the norm ball. Hence, the determination of adversarial example is completely based on the classification. This is because, as suggested before, the test cases are only generated from those SMILES strings with the same molecular structure.

\begin{table}[t]
\centering
\caption{Summary of RNN models under testing} 
\resizebox{\linewidth}{!}
{
\begin{tabular}{ccccc}
\hline
Test Model & No. of Classes & Test Acc. & Seq. of Interest & oracle \\ \hline
MNIST & 9 & 98.7\% & {[}4,24{]} & 0.01 \\
IMDB & 2 & 86.2\% & {[}400,500{]} & 0.05 \\
Lipophilicity & None & RMSE = 0.6278 & {[}60,80{]} & None \\
UCF101 & 101 & 88.6\% & {[}1,11{]} & 0.1 \\ \hline
\end{tabular}
\label{table:model}
}
\end{table}

\subsection{Diversity of Test Cases}\label{sec:diversityexp}

\begin{table}[t]
\centering
\caption{Configuration of test metrics} 
\resizebox{\linewidth}{!}
{
\begin{tabular}{cc}
\hline
Coverage Metrcis & Parameter Configuration \\ \hline
Neuron Coverage (NC) & Threshold = $0$ \\
K-multisection Neuron Coverage (KMNC) & $k = 10$ \\
Neuron Boundary Coverage (NBC) & LB = $-0.7$, UB = $0.7$ \\
Strong Neuron Activation Coverage (SNAC) & UB = $0.7$ \\
Boundary Coverage (BC) & $\alpha_{max} = 0.8$ \\
Step-wise Coverage (SC) & $\alpha_{sc} = 0.6$ \\
Temporal Coverage (TC) & $w = 5$, $|\Gamma|=3$ \\ \hline
\end{tabular}
}
\label{tab:parameters}
\end{table}

Test metrics can be seen as a proxy to exploit the input space, and intuitively more diversified test cases will explore more input space and thus are more likely to uncover different defects. Thus, we investigate if the achievement of high coverage will indeed lead to the testing of different LSTM functions (\textbf{RQ1}), and if our new metrics encourage the exploitation of more regions in the input space than existing metrics (\textbf{RQ2}). 

\subsubsection{Approximation of LSTM functional coverage (\textbf{RQ1})}
\label{sec:function-coverage}
Table \ref{table:motivation} shows that LSTM model's functional coverage can be approximated by using \testRNN metrics. This is based on the assumption that a data label (i.e., category) corresponds to a ``functional feature'' of the LSTM. We observe that, by only using one category of seeds input, it is hard to achieve high coverage rate for \testRNN metrics, even when thousands of test cases are generated. In contrast, with seeds input from more categories, the generated test cases from targeted mutation can broadly explore the input space and more internal behaviours of RNNs. Thus, all rates of \testRNN coverage metrics are significantly improved, given the test set. Table \ref{table:motivation} also records the coverage of neuron level metrics, which are widely used in the CNNs/FNNs. These test metrics show less sensitivity with respect to the diversity of functional features in the test suite, e.g., the NC coverage can already reach almost 100\% by only using test cases of one label.
 
 \begin{tcolorbox}
 \textbf{Answer to RQ1:}  The exploitation of internal behaviour by \testRNN can approximate the testing of different LSTM functional features.
 \end{tcolorbox}

\begin{table}[ht]

\centering
\caption{Impact of seeds to coverage metrics} 
\resizebox{\linewidth}{!}
{
\begin{tabular}{cccccccccc}
\hline
\multirow{2}{*}{Test Model} & \multirow{2}{*}{\begin{tabular}[c]{@{}c@{}}Seeds Input\\  \& Test Cases\end{tabular}} & \multirow{2}{*}{\begin{tabular}[c]{@{}c@{}}Categories\\ of Seeds\end{tabular}} & \multicolumn{7}{c}{Coverage Metrics} \\
 &  &  & NC & KMNC & NBC & SNAC & BC & SC & TC \\ \hline
\multirow{2}{*}{MNIST} & \multirow{2}{*}{100 / 5000} & 1 & 0.93 & 0.65 & 0.41 & 0.44 & 0.10 & 0.43 & 0.38 \\
 &  & 10 & \textbf{1.00} & \textbf{0.88} & \textbf{0.77} & \textbf{0.80} & \textbf{0.95} & \textbf{0.86} & \textbf{0.79} \\ \hline
\multirow{2}{*}{IMDB} & \multirow{2}{*}{100 / 5000} & 1 & 1.00 & 0.29 & 0.01 & 0.01 & 0.23 & 0.45 & 0.24 \\
 &  & 2 & \textbf{1.00} & \textbf{0.37} & \textbf{0.01} & \textbf{0.01} & \textbf{0.81} & \textbf{0.54} & \textbf{0.64} \\ \hline
\multirow{2}{*}{Lipophilicity} & \multirow{2}{*}{10 / 2000} & 1 & 0.81 & 0.31 & 0.05 & 0.06 & 0.00 & 0.00 & 0.04 \\
 &  & 10 & \textbf{1.00} & \textbf{0.86} & \textbf{0.68} & \textbf{0.66} & \textbf{0.95} & \textbf{0.95} & \textbf{0.90} \\ \hline
\multirow{2}{*}{UCF101} & \multirow{2}{*}{100 / 5000} & 1 & 1.00 & 0.47 & 0.07 & 0.06 &  0.00& 0.00 & 0.16 \\
 &  & 10 & \textbf{1.00} & \textbf{0.76} & \textbf{0.36} & \textbf{0.34} & \textbf{0.58} & \textbf{0.58} & \textbf{0.67} \\ \hline
\end{tabular}
\label{table:motivation}
}
\end{table}

\subsubsection{Comparison with Neuron Level Coverage (\textbf{RQ2})}
\label{sec:existing-coverage}
We implement the Neuron Coverage (NC) \cite{pei2017towards}, k-multisection Neuron Coverage (KMNC), Neuron Boundary Coverage (NBC) and Strong Neuron Activation Coverage (SNAC) \cite{ma2018deepgauge} 
%which are widely used in testing CNNs, 
on the testing layers of our LSTM models. We note that the concept ``neuron'' is ambiguous in RNNs, since the hidden output of RNNs' cells are vectors. Here, we consider covering each element of the hidden output $h$ in the testing layer. Results are presented in Table \ref{table:motivation} and \ref{table:complementarity}.

\begin{table}[ht]
\centering
\caption{Complementarity of test metrics: comparison between neuron level test metrics and the proposed \testRNN metrics in minimal test suite} 
\resizebox{\linewidth}{!}
{
\begin{tabular}{ccccccccc}
\hline
\multirow{2}{*}{Test Model} & \multirow{2}{*}{Target Metrics} & \multicolumn{4}{c}{Neuron Level Metrics} & \multicolumn{3}{c}{\testRNN Metrics} \\
 &  & NC & KMNC & NBC & SNAC & BC & SC & TC \\ \hline
\multirow{4}{*}{MNIST} & NC & \textbf{1.00} & 0.61 & 0.39 & 0.44 & 0.10 & 0.00 & 0.10 \\
 & KMNC & 1.00 & \textbf{0.85} & 0.67 & 0.72 & 0.10 & 0.29 & 0.44 \\
 & NBC & 1.00 & 0.77 & \textbf{0.73} & 0.75 & 0.14 & 0.19 & 0.28 \\
 & SNAC & 0.98 & 0.73 & 0.62 & \textbf{0.75} & 0.10 & 0.14 & 0.19 \\ \hline
\multirow{4}{*}{IMDB} & NC & \textbf{1.00} & 0.23 & 0.00 & 0.00 & 0.00 & 0.00 & 0.02 \\
 & KMNC & 1.00 & \textbf{0.39} & 0.01 & 0.01 & 0.01 & 0.02 & 0.07 \\
 & NBC & 0.48 & 0.13 & \textbf{0.01} & 0.01 & 0.00 & 0.00 & 0.01 \\
 & SNAC & 0.47 & 0.10 & 0.01 & \textbf{0.01} & 0.00 & 0.00 & 0.00 \\ \hline
\multirow{4}{*}{Lipophilicity} & NC & \textbf{1.00} & 0.43 & 0.16 & 0.16 & 0.05 & 0.05 & 0.03 \\
 & KMNC & 1.00 & \textbf{0.92} & 0.70 & 0.69 & 0.40 & 0.20 & 0.38 \\
 & NBC & 1.00 & 0.84 & \textbf{0.81} & 0.78 & 0.50 & 0.20 & 0.22 \\
 & SNAC & 1.00 & 0.77 & 0.62 & \textbf{0.78} & 0.40 & 0.20 & 0.13 \\ \hline
\multirow{4}{*}{UCF101} & NC & \textbf{1.00} & 0.48 & 0.26  & 0.36 & 0.10 & 0.10 & 0.15 \\
 & KMNC & 1.00 & \textbf{0.74} & 0.60 & 0.66 & 0.18 & 0.10 & 0.20 \\
 & NBC & 1.00 & 0.65 & \textbf{0.75} & 0.58 & 0.15 & 0.18 & 0.16 \\
 & SNAC & 1.00 & 0.76 & 0.68 & \textbf{0.84} & 0.22 & 0.18 & 0.20 \\ \hline
\end{tabular}
\label{table:complementarity}
}
\end{table}

In the experiments, we find that NC can be trivially achieved. Shown in Table \ref{table:motivation}, NC is not suitable for exploring RNNs' internal functionality, since one category's seeds input is enough for the high coverage of neuron activation. Moreover, we find that other neuron level test metrics may be impossible to satisfy for IMDB test model. The low coverage rate of KMNC, NBC and SNAC indicates that the activation of neurons for IMDB model is concentrated in a small interval. In other words, the neuron level test metrics cannot be a good option to search for diverse test cases. 

Table \ref{table:complementarity} shows the complementarity of neuron level test metrics and our proposed \testRNN metrics. A set of complementary test metrics (and test cases) can enhance the diversity of the testing. In the experiments, we take \textbf{minimal test suite}, in which the removal of any test case may lead to the reduction of coverage rate. The consideration of the minimality of test suite enables a fair comparison since it reduces the overlaps as much as possible. The results confirm that a test suite which can achieve high coverage for neuron level test metrics is not necessary to get the high coverage for RNN test metrics. For example, in the MNIST LSTM model, test cases that achieve 100\% NC can only cover less than 10\% of the overall test conditions by \testRNN metrics (with 10\% BC, 0\% SC and 10\% TC). Similar patterns also happen to other models.
That means the proposed test metrics provide the guide for the selection of additional test cases, which are complementary to those guided by the neuron level test metrics.

Moreover, we discover that there are many redundancy of test requirements, regarding to the relation between individual test metric in neuron level category. For example, if we derive a test suite which targets at increasing the coverage of KMNC, NBC and SNAC both get the high coverage results.

 \begin{tcolorbox}
 \textbf{Answer to RQ2:}  The \testRNN metrics exhibit a dramatic portion of LSTM internal behaviours that cannot be explored by existing metrics.
 \end{tcolorbox}

\subsection{Detecting RNN Defects}\label{sec:detectionexp}

\subsubsection{Searching for Adversarial Samples (\textbf{RQ3})}
\label{sec:adversarial-examples}

\begin{figure}[ht]
\centering
\includegraphics[width=\linewidth]{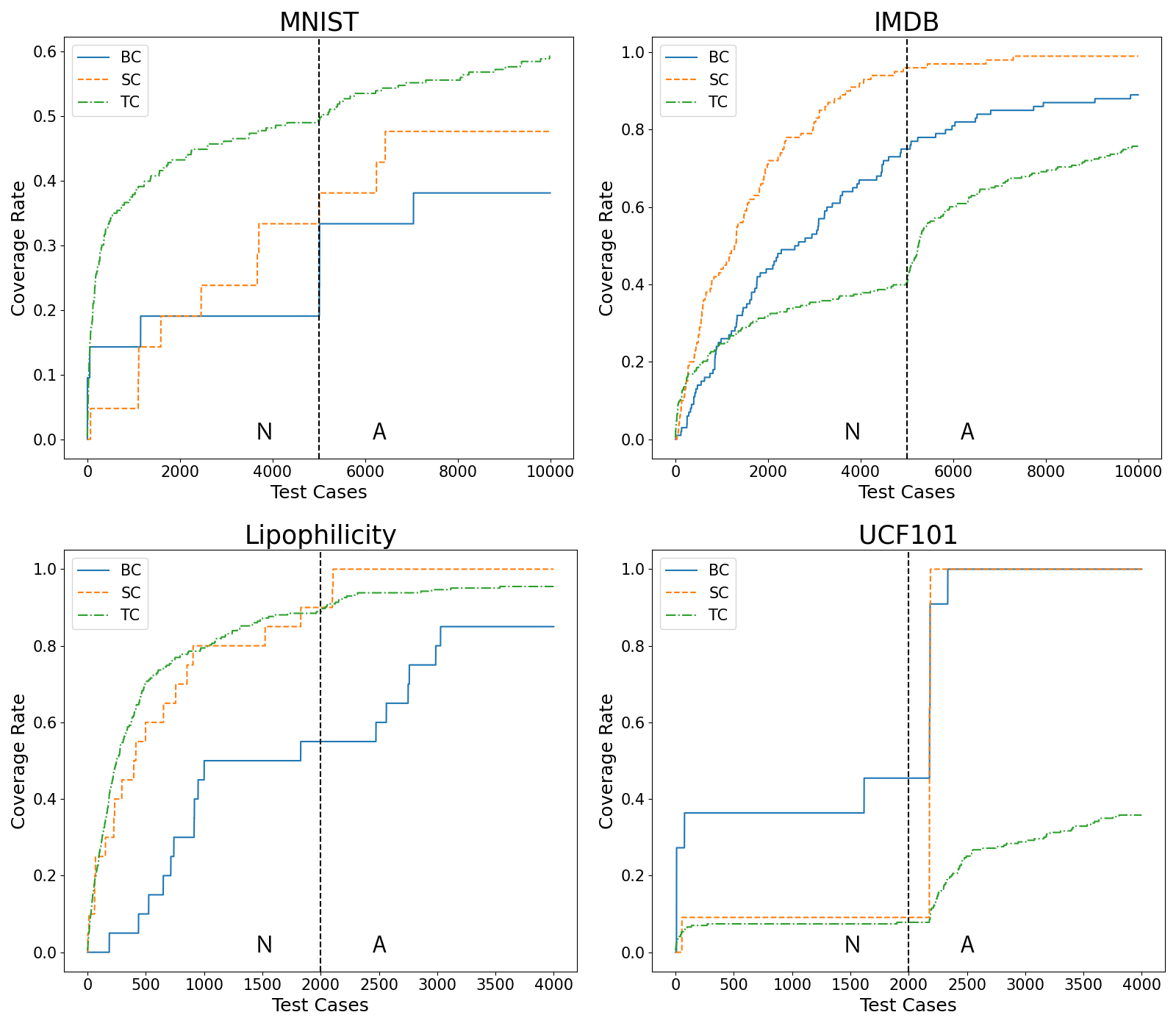}
\caption{Update of coverage with normal perturbed samples (`N') and adversarial samples (`A')}
\label{fig:coverage_update}
\end{figure}

We collect the set of normal perturbed samples (N) and adversarial samples (A), respectively. First, normal perturbed samples are added to the test set  to witness the increase of the coverage. When the coverage is difficult to improve, adversarial samples are considered. The update of whole process is illustrated in Fig.~\ref{fig:coverage_update}. The dashed vertical line distinguish the coverage update with normal perturbed samples from that with adversarial samples. It should be noted that the coverage update of some test metrics is stepped growth, due to the small amount of total test conditions which is shwon in Table \ref{table:model}. 

Fig.~\ref{fig:coverage_update} reveals that normal perturbed samples can only satisfy part of test conditions, while the rest are more sensitive to the adversarial samples. In all the plots, coverage of RNN test metrics can be further increased in consideration of adversarial samples. A more obvious example is, the TC coverage of IMDB model tend to saturate in the left side when only normal perturbed samples are utilized. In the right side, the coverage curve becomes steep, indicating the discovery of test cases capturing new internal behaviors.

In addition to the sensitivity of test metrics to adversarial samples, we show how to compare the robustness of models via coverage guided testing. We use TC as the termination condition to generate a test suite and calculate the adversarial rate. To achieve the high coverage of test metrics, we use genetic algorithm for test case generation, more details of which can be seen in Section \ref{sec:mutation}. The other settings remain the same for the fair comparison. 

\begin{table}[ht]
\centering
\caption{Comparing the robustness of models via coverage guided testing} 
\resizebox{\linewidth}{!}
{
\begin{tabular}{cccccccc}
\hline
\multirow{2}{*}{Test Dataset} & \multirow{2}{*}{\begin{tabular}[c]{@{}c@{}}Model\\ No.\end{tabular}} & \multirow{2}{*}{Test Cases} & \multirow{2}{*}{\begin{tabular}[c]{@{}c@{}}Adv. Samples\\ Rate\end{tabular}} & \multirow{2}{*}{\begin{tabular}[c]{@{}c@{}}Unique\\ Adv. Samples\end{tabular}} & \multicolumn{3}{c}{Coverage Metrics} \\  
 &  &  &  &  & BC & SC & TC \\ \hline
\multirow{2}{*}{MNIST} & 1 & 5958 & \textbf{0.060} & \textbf{176} & 0.48 & 0.81 & 0.90 \\
 & 2 & 3570 & 0.075 & 184 & 0.57 & 0.86 & 0.90 \\ \hline
\multirow{2}{*}{IMDB} & 1 & 5841 & \textbf{0.039} & 138 & 0.94 & 0.93 & 0.75 \\
 & 2 & 1575 & 0.047 & \textbf{68} & 0.62 & 0.72 & 0.75 \\ \hline
\multirow{2}{*}{Lipophilicity} & 1 & 2936 & 0.371 & 191 & 0.95 & 1.00 & 0.95 \\
 & 2 & 6727 & \textbf{0.010} & \textbf{44} & 0.88 & 1.00 & 0.95 \\ \hline
\multirow{2}{*}{UCF101} & 1 & 6352 & 0.420 & 182 & 0.98 & 0.95 & 0.60 \\
 & 2 & 6100 & \textbf{0.250} & \textbf{90} & 0.95 & 0.92 & 0.60 \\ \hline
\end{tabular}
\label{table:adv_test}
}
\end{table}

As shown in Table \ref{table:adv_test}, adversarial samples rate and number of unique adversarial samples in the generated test suite are two important indicators for the robustness evaluation. The unique adversarial samples refer to the adversarial samples crafted from distinct seeds input. For a set of trained models, we pursue the model, the test suite of which contains less amount of adversarial samples and unique adversarial samples. For example, we pick up model 2 for ipophilicity prediction, since the values of two indicators are way smaller than that of model 1. We comment that with large enough amount of test cases, coverage-guide testing approach provides a new way for the measure and selection of more robust classifier.  This is compatible with the results in \cite{sun2018testing} that a poorly trained neural network exposes more adversarial samples subject to well-defined coverage guided testing.  

 \begin{tcolorbox}
 \textbf{Answer to RQ3:} By exploiting the model's internal behaviours, \testRNN is able to capture the LSTM adversarial samples. 
 \end{tcolorbox}

\subsubsection{Detecting Backdoor input in RNNs (\textbf{RQ4})}
\label{sec:backdoor}

We investigate the possibility of applying coverage-guided testing to the detection of backdoor input in neural networks. We try to exploit if there is any difference between clean input and backdoor input which can be captured by our proposed test metrics. Examples of backdoor input are illustrated in Fig.~\ref{fig:adv}. We train two handwritten digits recognition models, one of which is benign classifier and the other one is the malicious classifier subject to the backdoor attack in \cite{gu2019badnets}. Table \ref{table:backdoor_test} shows that, both benign and malicious classifiers keep good prediction performance in clean test set. For the backdoor test set, benign classifier keep the normal accuracy, while the malicious classifier predicts inputs with the backdoor trigger as the attacked label successfully.  

\begin{table}[ht]
\centering
\caption{Sensitivity of test metrics to backdoor samples in MNIST dataset} 
\resizebox{\linewidth}{!}
{
\begin{tabular}{cccccccccccc}
\hline
\multirow{2}{*}{Model} & \multirow{2}{*}{\begin{tabular}[c]{@{}c@{}}Test Acc.\\ (C / B)\end{tabular}} & \multirow{2}{*}{Data} & \multicolumn{3}{c}{Class 0} & \multicolumn{3}{c}{Class 6} & \multicolumn{3}{c}{Class 9} \\
 &  &  & BC & SC & TC & BC & SC & TC & BC & SC & TC \\ \hline
\multirow{3}{*}{Benign} & \multirow{3}{*}{99.1\% / 9.5\%} & T & 0.39 & 0.25 & 0.16 & 0.29 & 0.18 & 0.30 & 0.32 & 0.29 & 0.22 \\
 &  & T + C & 0.39 & 0.25 & 0.17 & 0.29 & 0.18 & 0.30 & 0.32 & 0.29 & 0.22 \\
 &  & T + B & 0.39 & 0.25 & 0.17 & 0.29 & 0.18 & 0.30 & 0.32 & 0.29 & 0.22 \\ \hline
\multirow{3}{*}{Malicious} & \multirow{3}{*}{98.7\% / 100\%} & T & 0.39 & 0.18 & 0.30 & 0.25 & 0.18 & 0.60 & 0.07 & 0.18 & 0.27 \\
 &  & T + C & 0.39 & 0.18 & 0.30 & 0.25 & 0.18 & 0.60 & 0.07 & 0.18 & 0.27 \\
 &  & T + B & 0.39 & \textbf{0.25} & \textbf{0.33} & 0.25 & \textbf{0.21} & \textbf{0.63} & 0.07 & \textbf{0.21} & \textbf{0.29} \\ \hline
\end{tabular}
\label{table:backdoor_test}
}
\end{table}

We conduct sensitivity analysis by computing the coverage of the proposed test metrics in training data (T), clean test data (C), and backdoor test data (B) for each classifier. In the first row of Table \ref{table:backdoor_test}, we calculate the coverage of the training data from same class. On the basis of this, we add clean test data or backdoor test data for evaluation. If the coverage rate is further increased in second and third row, the new internal patterns are discovered. The experimental results describe that backdoor input activate same internal behavior with clean input for a benign classifier. In contrast to this, the backdoor input to malicious classifier will induce different internal activation, which can be seen from the apparent increase of coverage in T+B. Although the backdoor input is very similar to the clean input with a small region of pixels changed (Fig.~\ref{fig:adv}), the internal activation in the malicious model can still be revealed by the coverage change of the proposed \testRNN metrics.

We remark that the above experiment only confirms that test metrics are sensitive to backdoor samples when testing an attacked model. More accurate detection of backdoor in RNNs needs more precise refinement of test metrics, e.g. adding the backdoor knowledge to the metrics design on top of the structure information. Nevertheless, the goal of coverage guided testing is still diversifying the test suite so that defects like backdoor samples are more likely to be detected.

 \begin{tcolorbox}
 \textbf{Answer to RQ4:}  The \testRNN metrics can identify the difference between the backdoor input and the normal input (to malicious models).
 \end{tcolorbox}

\subsection{Effectiveness of Test Case Generation (\textbf{RQ5})} \label{sec:test_case_generationexp}

\begin{figure}[ht]
    \centering
    \includegraphics[width=\linewidth]{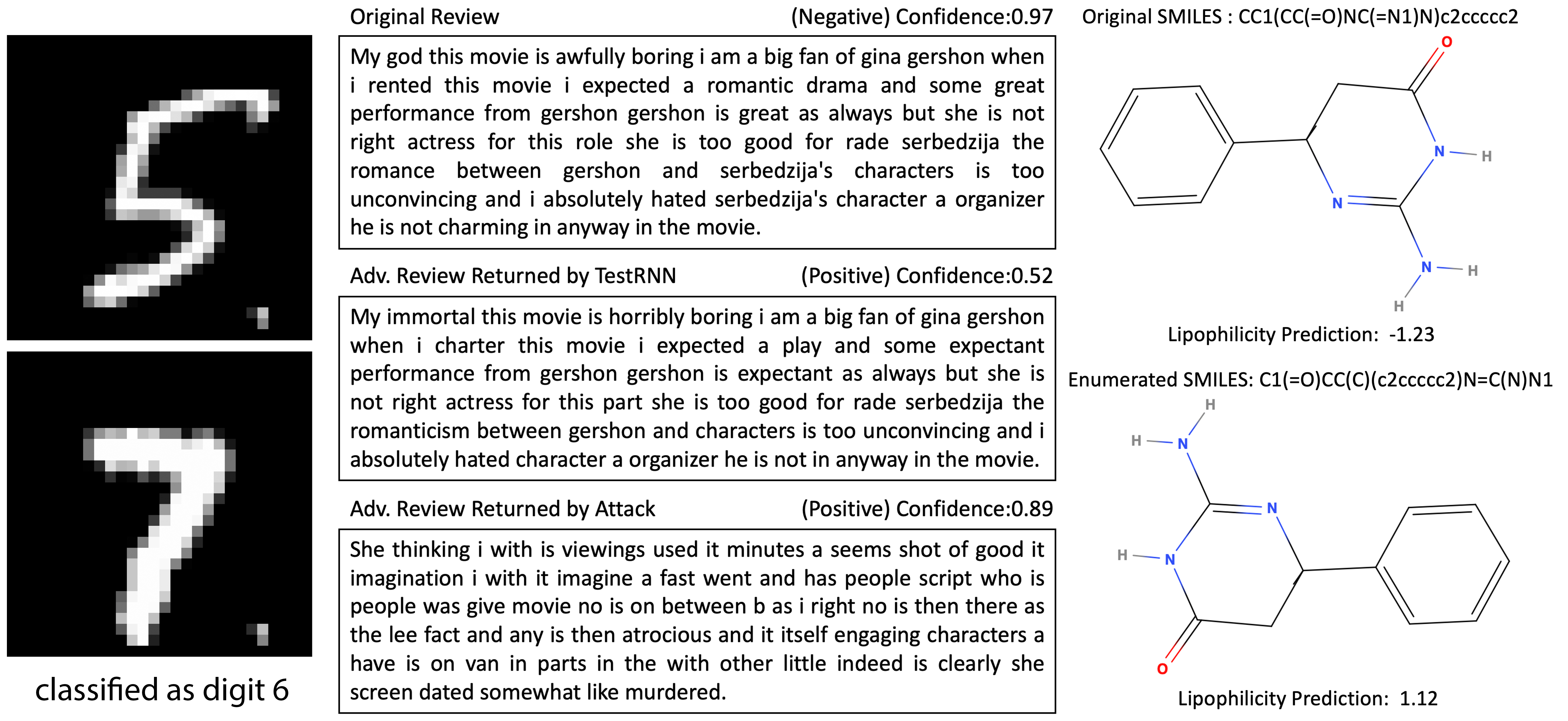}
    \caption{Backdoor samples for MNIST model (left). Adversarial samples for IMDB (middle) and Lipophilicity (right) models.}
    \label{fig:adv}
\end{figure}

We show the effectiveness of our test case generation from the following aspects: (1) it is non-trivial to achieve high coverage rate, and (2) there is a significant percentage of adversarial samples in the generated test suite. For (1), we show that the targeted mutation (i.e., random mutation enhanced by genetic algorithm) is needed to boost the coverage rate. 
Three test case generation methods are considered: (Seeds) sampling 200 seeds input 
from training dataset, (Ran.) generating test cases from seeds by using random mutation, and (Targ.) generating test cases from seeds by using targeted mutation. 
Fig.~\ref{fig:adv} demonstrates detected adversarial samples for IMDB and Lipophilicity models, and we omit other models for brevity.
All experimental results are based on 5 runs with different random seeds. The results are averaged and summarised in Table~\ref{tab:thresholdexperiments}. For each test case generation method and LSTM model, we also report the number of adversarial samples, unique adversarial samples in the test suite and their average perturbation. This experiment considers all four models. 

Table~\ref{tab:thresholdexperiments} shows that, the coverage rates and the number of adversarial samples for Ran. are 
significantly higher than those of Seeds, that is, Ran. is effective in finding the adversarial samples around the original seeds.
Furthermore, if we use Targ., both the coverage rates and the number of adversarial samples are further increased. 
%In the meantime, the number of adversarial samples also experiences a non-trivial growth. 
The above observations confirm the following two points: (1) our test metrics come with a strong bug finding ability; and  (2) higher coverage rates indicate more comprehensive test. We remark that, the TC rates for UCF101 model are relatively low and harder to improve, because the mutations are made on the image frames (i.e., before CNN layers) instead of directly on the LSTM input. This shows that the adversarial samples for CNNs are orthogonal to those of LSTMs, another evidence showing that test metrics for CNNs cannot be directly applied to RNNs. 
 \begin{tcolorbox}
 \textbf{Answer to RQ5:} The test case generation algorithm is effective in improving both the coverage rate and the adversary rate. In particular, the targeted mutation method can be utilised to find more corner samples.
 \end{tcolorbox}

\begin{table}[ht]
\centering
\caption{Experiments for Test Case Generation Methods}
\resizebox{\linewidth}{!}
{
\begin{tabular}{ccccccccc}
\hline
\multirow{2}{*}{Test Model} & \multirow{2}{*}{\begin{tabular}[c]{@{}c@{}}Test Gen.\\ Method\end{tabular}} & \multirow{2}{*}{Test Cases} & \multirow{2}{*}{\begin{tabular}[c]{@{}c@{}}No. of Adv.\\ samples\end{tabular}} & \multirow{2}{*}{\begin{tabular}[c]{@{}c@{}}Avg. Perturb.\end{tabular}} & \multirow{2}{*}{\begin{tabular}[c]{@{}c@{}}Unique\\ Adv. Samples\end{tabular}} & \multicolumn{3}{c}{Coverage Metrics} \\
 &  &  &  &  &  & BC & SC & TC \\ \hline
\multirow{3}{*}{MNIST} & Seeds & 200 & - & - & - & 0.43 & 0.14 & 0.34 \\
 & Ran. & 10000 & 226 & 1.180 & 18 & 0.57 & 0.52 & 0.66 \\
 & Targ. & 10000 & \textbf{244} & \textbf{1.497} & \textbf{32} & \textbf{1.00} & \textbf{1.00} & \textbf{0.79} \\ \hline
\multirow{3}{*}{IMDB} & Seeds & 200 & - & - & - & 0.11 & 0.05 & 0.24 \\
 & Ran. & 10000 & 308 & 0.136 & 88 & 0.84 & 0.40 & 0.77 \\
 & Targ. & 10000 & \textbf{367} & \textbf{0.103} & \textbf{97} & \textbf{1.00} & \textbf{0.58} & \textbf{0.82} \\ \hline
\multirow{3}{*}{Lipophilicity} & Seeds & 200 & - & - & - & 0.65 & 0.55 & 0.48 \\
 & Ran. & 2000 & 812 & - & 190 & 0.95 & 1.00 & 0.91 \\
 & Targ. & 2000 & \textbf{834} & - & \textbf{194} & \textbf{1.00} & \textbf{1.00} & \textbf{0.95} \\ \hline
\multirow{3}{*}{UCF101} & Seeds & 200 & - & - & - & 0.52 & 0.53 & 0.11 \\
 & Ran. & 10000 & 3613 & 1.031 & 112 & 0.82 & 0.90 & 0.31 \\
 & Targ. & 10000 & \textbf{4201} & \textbf{1.251} & \textbf{156} & \textbf{1.00} & \textbf{1.00} & \textbf{0.66} \\ \hline
\end{tabular}
\label{tab:thresholdexperiments}
}
\end{table}

\subsection{Comparison with Attack-based Defect Detection (\textbf{RQ6})}
\label{sec:comparison_attack}

We compare \testRNN with state-of-the-art RNN adversarial attack \cite{papernot2016crafting,behjati2019universal}, which detects robustness defects. These attack algorithms utilise the model's gradient over input sequence to iteratively change some parts of the input that contribute the most to the model's prediction. Their methods can successfully find adversarial samples. However, these attack methods have two main drawbacks, when compared with our testing method.

First, attack methods search for adversarial samples by adding perturbations in the gradient direction. This easily leads to the situation where the generated adversarial samples are concentrated in a ``buggy'' area of the input space, as shown in Table~\ref{tab:diversity_measurement} and Fig.~\ref{fig:attack_testrnn_deepstellar_mnist}. We first collect the same amount of adversarial samples in MNIST model returned by attack methods and \testRNN, respectively. Then, we calculate the angular-based diversity of each set (Table~\ref{tab:diversity_measurement}) and apply the Principal Component Analysis (PCA), a well known dimensionality reduction technique, to project the high dimensional adversarial images onto two dimensional space for better visualisation (Fig.~\ref{fig:attack_testrnn_deepstellar_mnist}). We can see from the resulting diversity measurement and visualisation that, compared to attack methods, our testing method exercises different behaviors of RNN and generates a diverse set of test cases, intensively covering the input region around the seed input. \emph{This ability will be helpful in exposing more types of defects of the RNN} (not merely in the gradient direction). 

Moreover, RNNs are widely applied to the nature language processing, in which the inputs to an RNN, i.e., words, are discretely distributed. Attack methods aggressively replace important words in the text and produce an adversarial sequence. In this process, it is hard to consider both the gradient and the whole text's semantic meaning. That is, the modified text may easily become human-unreadable and impossible to occur in real world. On the other hand, our testing method is able to reduce such problems by taking the mutants from off-the-shelf tools such as the EDA toolkit. Fig.~\ref{fig:adv} presents adversarial movie reviews returned by attack method and \testRNN, respectively.  It is easy to see that the adversarial review returned by the gradient attack is hard to comprehend while the one from \testRNN\ is much easier. 

 \begin{tcolorbox}
 \textbf{Answer to RQ6:} The \testRNN is able to generate a set of diverse and natural test cases, so as to expose more types of defects.
 \end{tcolorbox}

\subsection{Comparison with State-of-the-Art testing methods (\textbf{RQ7})}\label{sec:compare-with-related-work}

We compare \testRNN with DeepStellar, a state-of-the-art testing tool dedicated for RNNs. As discussed in Section \ref{sec:comparison_metrics}, two different test metrics are integrated in DeepStellar, i.e. state coverage and transition coverage, which are corresponding to boundary coverage and step-wise coverage in \testRNN. Apart from these, \testRNN\ have temporal coverage for the internal sequential processing behaviour of RNNs. We start from 100 seeds drawn from training set and generate 100000 test cases by DeepStellar and \testRNN, respectively. The test suites are evaluated for the coverage rate and number of adversarial samples. We compute basic state coverage (BS), basic transition coverage (BT), and weighted transition coverage (WT) in DeepStellar guided by different generation strategy, S-Guide and T-Guide. The testing results for both are recorded in Table \ref{tab:related_work_1}. First, we can see that test metrics in DeepStellar already have high coverage rates upon seeds input, as opposed to our metrics which display relatively smaller coverage rates upon the same seeds. That means that our metrics are better for exploiting the input space around seeds. Second, DeepStellar adopts the fuzzing strategy with the guidance of different test metrics, which is effective to boost the coverage. However, in this experiment for small-scale model trained on MNIST, 100000 test cases are still not enough for 100\% coverage of test requirements in DeepStellar. It seems that some of their defined test requirements may be infeasible to satisfy. On the contrary, \testRNN can achieve a relatively high coverage results with random mutation and the coverage rates of all the metrics can be significantly boost to achieve 100\% by genetic algorithm based mutation method. The number of adversarial samples in the test suite reflects that \testRNN\ is superior to DeepStellar in terms of exploiting diverse internal behaviors and bugs finding ability.

\begin{table}[ht]
\centering
\caption{Comparison between DeepStellar and \testRNN using MNIST: 100000 test cases are generated from 100 seeds}
\resizebox{\linewidth}{!}
{
\begin{tabular}{cccccccc}
\hline
\multicolumn{4}{c}{DeepStellar} & \multicolumn{4}{c}{TestRNN} \\
Test Metrics & Seeds & S-Guid. & T-Guid. & Test Metrics & Seeds & Ran. & Targ. \\ \hline
BS & 0.45 & 0.80 & 0.82 & BC & 0.14 & 0.57 & 1.00 \\
BT & 0.11 & 0.32 & 0.63 & SC & 0.38 & 0.67 & 1.00 \\
WT & 0.76 & 0.90 & 0.95 & TC & 0.24 & 0.70 & 1.00 \\
Adv. Samples & - & 1588 & 1661 & Adv. Samples & - & 1778 & 1830 \\ \hline
\end{tabular}
\label{tab:related_work_1}
}
\end{table}

\begin{table}[ht]
\centering
\caption{Angular-based diversity (a greater value represents a better diversity) and average perturbation (smaller is better) of adversarial samples}
\resizebox{\linewidth}{!}
{
\begin{tabular}{ccccccc}
\hline
\multirow{2}{*}{Seed} & \multicolumn{3}{c}{Angular-based Diversity} & \multicolumn{3}{c}{Avg. Perturb.} \\
 & \testRNN & DeepStellar & Attack & \testRNN & DeepStellar & Attack \\ \hline
1 & -0.277 & -0.468 & -0.598 & 0.006 & 0.012 & 0.014 \\
2 & -0.289 & -0.438 & -0.556 & 0.006 & 0.010 & 0.013 \\ \hline
\end{tabular}
\label{tab:diversity_measurement}
}
\end{table}

\begin{figure}[ht]
\centering
\includegraphics[width=\linewidth]{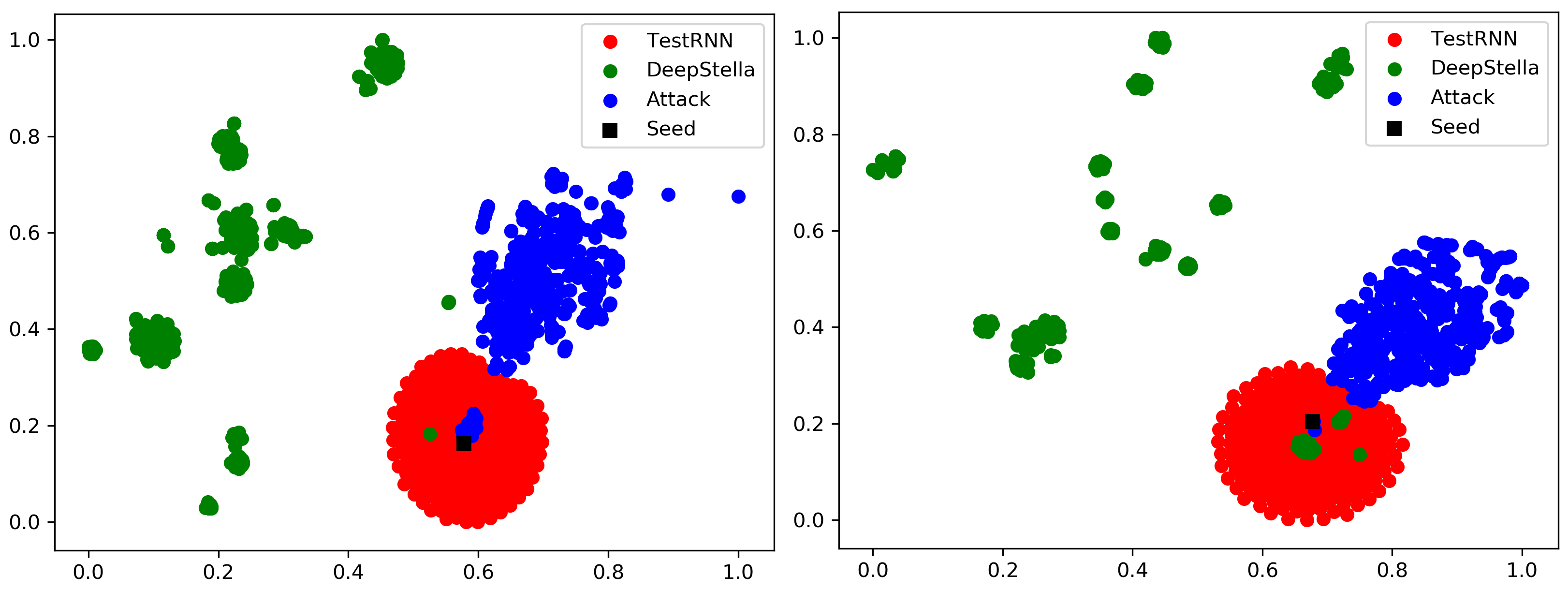}
\caption{Visualisation of adversarial samples generated by \testRNN, DeepStellar, and Gradient-based Attack, respectively, in MNIST model. The visualisation is conducted by projecting high-dimensional images onto a two-dimensional space. Each figure corresponds to a seed input in the dataset. }
\label{fig:attack_testrnn_deepstellar_mnist}
\end{figure}

To understand the relative merits of the defects returned by DeepStellar, \testRNN, and Gradient-based Attack, respectively, we compute the angular-based diversity and average perturbation of each set (Table~\ref{tab:diversity_measurement}), and visualise them with PCA projection (Fig.~\ref{fig:attack_testrnn_deepstellar_mnist}). We can see that the adversarial samples from DeepStellar are sparsely distributed and most of them are more distant to the seed input. \testRNN explores space that is close to the seed input. This aligns better to the goal of adversarial testing, which is to find more bugs around the seed with as small perturbations as possible (the bugs are more realistic/natural, thus more likely to exist in real world).

\begin{table}[ht]
\centering
\caption{Complementarity of test metrics in DeepStellar and \testRNN}
\resizebox{\linewidth}{!}
{
\begin{tabular}{ccccccccc}
\hline
\multirow{2}{*}{Tool} & \multirow{2}{*}{\begin{tabular}[c]{@{}c@{}}Test Gen.\\ Method\end{tabular}} & \multirow{2}{*}{\begin{tabular}[c]{@{}c@{}}Target\\ Metrics\end{tabular}} & \multicolumn{6}{c}{Coverage Metrics} \\
 &  &  & BS & BT & WT & BC & SC & TC \\ \hline
\multirow{2}{*}{TestRNN} & Ran. & - & 0.78 & 0.63 & 0.94 & 0.57 & 0.67 & 0.70 \\
 & Targ. & BC,SC,TC & 0.78 & 0.64 & 0.95 & 1.00 & 1.00 & 1.00 \\
\multirow{2}{*}{DeepStellar} & S-Guide. & BS & 0.80 & 0.32 & 0.90 & 0.05 & 0.10 & 0.12 \\
 & T-Guide. & BT,WT & 0.82 & 0.63 & 0.95 & 0.10 & 0.24 & 0.20 \\ \hline
\end{tabular}
\label{tab:related_work_2}
}
\end{table}

In addition to the comparison of testing results, we are also interested in the complementarity of test metrics in \testRNN\ and DeepStellar. We derive the minimal test suites by different test case generation methods. Then, the test suite generated by \testRNN\ is evaluated for the coverage of metrics in DeepStellar, and vice versa. Results in Table \ref{tab:related_work_2} suggests that test suite generated by \testRNN\ can easily achieve high coverage rate for the metrics in DeepStellar. We find that, test suite produced by DeepStellar cannot get high coverage rate on our metrics. This confirms our discussion of the relation between coverage metrics in Section \ref{sec:comparison_metrics}.  
 \begin{tcolorbox}
 \textbf{Answer to RQ7:} The \testRNN test generation can achieve high coverage of the test metrics in DeepStellar, but not vice versa. 
 \end{tcolorbox}

\subsection{Exhibition of Internal Working Mechanism (\textbf{RQ8})}
\label{sec:interpretable}

In this section, we show that the working mechanism of LSTM networks can be understood via the test cases generated from \testRNN. We conduct experiments to visualise the learning process of LSTM layer via \testRNN results.

\begin{figure}[ht]
\centering
\includegraphics[width=\linewidth]{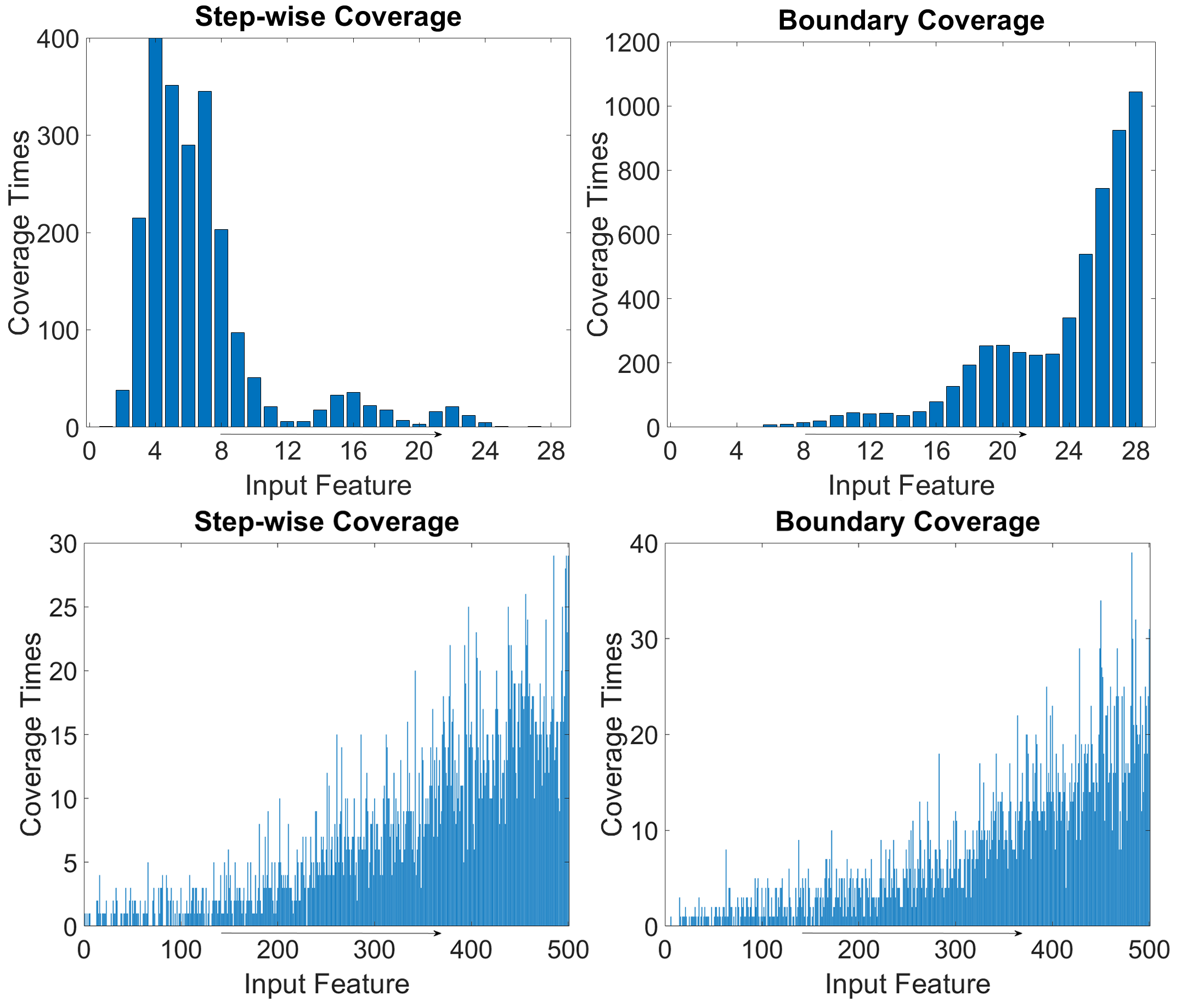}
\caption{2000 test cases are used to demonstrate the coverage times of 28 features in an LSTM layer of MNIST model (first line) and 500 input features in LSTM layer of IMDB Sentiment Analysis model (second line).}
\label{fig:coverage_times}
\end{figure}

\textbf{Coverage times} denote the number of times a test condition is satisfied by running the test suite. Intuitively, coverage times represent the level of difficulty of asserting an input feature.  Fig.~\ref{fig:coverage_times} reports the coverage times for each input feature. We note that, in BC and SC, each input feature $x_t$ corresponds to a test condition on $\xi_{t}^{s,a}$, as in MNIST it is defined with respect to a row of pixels on the input image. In sentiment analysis model, the input feature refers to a word in movie reviews.

As discussed in Section~\ref{sec:metrics}, SC is to assert if an input feature is significant to the model prediction. Then an important input feature will cause great changes of hidden memory $h_t$ and satisfy the test condition of SC. BC monitors the forget gate values at each time step. The satisfaction of BC means the LSTM will not drop out the information stored in memory.

If we combine SC and BC plots, the whole working process of LSTM layer inside the MNIST model becomes transparent. The sequential input of an image starts from the top row and finishes at the bottom row. At the beginning, the top rows of MNIST images are blank and do not contribute to the model prediction. These less-important information is gradually thrown away from the memory. When the input rows containing digits are fed to the LSTM cells, the model will start learning and the short term memory, represented by the outputs $h_t$, start to have strong reactions. 
When approaching the end of the input, which corresponds to the bottom of the digit images, LSTM has already been confident about the final classification and therefore becomes lazy to update the memory. Overall, we can see that, MNIST digits recognition is not a complicated task for the LSTM model and usually the top half of the images are sufficient for the classification.

For the IMDB model, the final classification is influenced by every input feature. To make sure that input features between 450-500 contain real words instead of padded 0s, we take 2,000 reviews whose length are greater than 50. We observe from the second line in Fig.~\ref{fig:coverage_times} that the coverage times gradually increase, it might be the nature of test cases -- most test cases contain text of length much less than 500. We therefore focus on the last 50 input features.
We see that, both BC and SC test conditions in the IMDB model are randomly activated, a phenomenon that is completely different from that of MNIST results. This can be explained as that the IMDB model does not have a fixed working pattern like the MNIST model. Sensitive words in a review may appear in any place of the text.

 \begin{tcolorbox}
 \textbf{Answer to RQ8:} The generated test suite can be used to understand the data processing mechanism of LSTM models. This is a step towards interpretable RNNs. 
 \end{tcolorbox}

\subsection{Threats to Validity}

First, we fix the thresholds or symbols of test metrics for all the experiments. If we decrease the values of threshold (or reduce the symbols to represent sub-ranges), the test conditions can be easier to satisfy, and fewer test cases are generated. Conversely, if we tighten the thresholds, more test cases are needed to cover the test conditions. The input space are more thoroughly explored.

 Second, we only choose part of input sequence to test, details of which is shown in Table \ref{table:model}. If we use \testRNN to test the entire input sequence, some test conditions may be harder to meet. The choices of partial input sequences in our experiment are as follows. For MNIST dataset, the hand-written digits are usually concentrated on 4th to 24th rows out of 28 rows. The rest of the images are blank. For IMDB and Lipophilicity dataset, the input to RNNs are usually padded with 0s. And the input 0s only induce very small activation, which can be seen in Fig.~\ref{fig:coverage_times}.

 We define unique mutation functions for different models to ensure the generated test input are always valid. Since we set thresholds of test metrics with reference to the training data, it is non-trivial to validate the test input. %The key point of 
 Mutation function needs to keep the semantics meanings of seeds input. For example, in the experiment of testing IMDB model, we mutate the text paragraph instead of the input to LSTM layer.
 
 Some minor threats include the settings of oracle and random seeds. We also fix the configurations for these parameters to make all the experiments consistent. The oracle radius can affect the adversarial samples rate and the average perturbations in the test suite. If we set up a smaller oracle radius, the number of perturbed input recognized as adversarial samples and the average perturbations are both decreased. The random seeds are utilized to control the reproducibility of the experiments. In most experiments, we do several test with different seeds input and get the average results so that the accidental errors can be avoided.

\section{Related Work}\label{sec:relatedworks}

\paragraph{Adversarial Samples for RNNs}

Since adversarial robustness is regarded as a major safety concern for deep learning \cite{2018arXiv181208342H}, a number of works appear on generating adversarial samples for RNN tasks such as natural language processing \cite{papernot2016crafting}, and 
automated speech recognition \cite{GP2017}. In this paper, we treat adversarial samples as a proxy to evaluate the effectiveness of the proposed coverage criteria.

\paragraph{Testing Feed-forward Neural Networks}

Most neural network testing methods focus on FNNs. In
\cite{PCYJ2017} the neuron coverage is proposed for exploiting neuron activation conditions in an FNN. Various refinements and extensions of neuron coverage are later developed in \cite{ma2018deepgauge}. Motivated by the usage of MC/DC coverage metrics in high criticality software, in \cite{sun2018testing}, a family of MC/DC variants are designed for FNNs, by taking into account the causal relation between features of different layers. Moreover, it has been shown in \cite{sun2018testing} that the criteria in \cite{PCYJ2017,ma2018deepgauge} are special cases of the MC/DC variants.

In addition to the structural coverage criteria mentioned above, metrics in  \cite{wicker2018feature,huang18b} define a set of test conditions to partition the input space. Though not being a  coverage metric, the method in \cite{kim2018guiding} measures the difference between training and test datasets
based on structural information of FNNs. 

Guided by the coverage metrics, test cases can be generated via various techniques including e.g., heuristic search \cite{deeproad,RWSHKK2019}, fuzzing \cite{odena2018tensorfuzz,dlfuzz}, mutation \cite{deepmutation,wang2018detecting}, and symbolic encoding \cite{sun2018concolic,gopinath2018symbolic}, etc. None of these works have considered RNNs. Please refer to \cite{2018arXiv181208342H} for a survey on techniques for the safety and trustworthiness of neural networks.

\paragraph{Testing RNNs}

Few works contribute to the development of coverage metrics for RNNs. DeepStellar \cite{du2019deepstellar} firstly proposes to abstract an RNN model into a Discrete-Time Markov Chain (DTMC). The abstracted DTMC is an approximation, whose fidelity to the original RNN is unknown. Such approximation can lead to unexpected consequences for testing, including the false positives and false negatives due to the misplacement of faulty corner cases in RNN and DTMC. Moreover, only cell states $c$ are utilised in the abstracted DTMC along with the development of test metrics. Other functional components of RNNs, including the gates $f,i,o$ and the hidden output $h$, are not considered. As demonstrated in experiments, these components have their dedicated meanings and ought to be considered when a more extensive testing is expected. They are also helpful to improve the interpretability of testing results. 
Moreover, RNN-Test \cite{guo2019rnn} develops some test metrics to work with the structural components (gate $f$, cell $c$, output $h$) directly. Their test metrics can be viewed as special cases of our boundary coverage. More importantly, they do not study the temporal relations, which we believe are the most fundamental characteristics of RNNs (as opposed to CNNs). We think that the differences mentioned above are significant enough to distinguish the work of ours from that in \cite{du2019deepstellar} and \cite{guo2019rnn}. 

\paragraph{Difference between Testing and Defect Detection}
Recent paper \cite{yan2020correlations} on correlations between coverage criteria and model quality suggests that coverage guided testing complements gradient-based adversarial attack. They discover that adversarial samples found by FNN coverage guided testing can be further utilised to retrain more robust models. However, such models may not be robust to the gradient-based attack (e.g. PGD \cite{madry2017towards}). On the other hand, PGD based adversarial training may improve models' robustness to the adversarial attack but not attacks with guidance of coverage metrics. 

\paragraph{Visualisation for LSTM}

In each LSTM layer, a sequential input $\{x_t\}_{t=1}^n$ corresponds with a sequence of vectors for structural components, e.g., $\{f_t\}_{t=1}^n$ and $\{h_t\}_{t=1}^n$. These internal vectors are high dimensional and impossible to be comprehended by humans. Then, some dimensionality reduction methods (e.g. t-SNE and PCA) have been adopted to visualise the information behind them. For instances, \cite{rauber2017visualizing} employs PCA to extract the principle component of $h_t$ at each time step $t$. These methods facilitate the interpretation of RNN's hidden behaviours. Dimensionality reduction methods have also been used to abstract a neural network into an abstracted model such as a Bayesian network \cite{berthier2021abstraction}. Our interpretation is completely different from the above, and works by visualising the working process of LSTM layer based on a set of test cases. 

\paragraph{Neural Network Repairing}
\label{sec:repair}

The repairing of neural network has also been studied, with the aim to utilise the generated test cases  to improve the model's adversarial robustness~\cite{ma2018mode,zhang2019apricot,robot2020} or fix the detected backdoor~\cite{neurospf2020}. In contrast to the typical machine learning re-training \cite{wang2019convergence}, such work often relies on properly designed test cases to first identify certain structures inside the neural network model that are responsible for the model's undesirable behaviours, and then correct the model's behaviour by e.g., re-training~\cite{ma2018mode}, weight adaption~\cite{zhang2019apricot}, symbolic constraint solving~\cite{neurospf2020}, etc. In \cite{robot2020}, each test case's impact on improving the model robustness is quantified.
Nevertheless, we are not aware of any repairing method that is designed for RNNs.

\section{Conclusions}\label{sec:concls}

In this paper, we propose a coverage guided test framework for the verification and validation of RNNs. We develop a tool \testRNN based on the test framework and validate it on several LSTM models, trained on popular benchmarks. In the future, we plan to investigate the possibility of utilising the testing results to mitigate the RNN defects and also certify RNNs.

\section*{Acknowledgement}
We thank the DeepStellar team for sharing their code and setup which make the comparison of experiments possible. We thank the anonymous reviewers of this manuscript for their inspiring comments that help us strengthen this work. This work is supported by the UK EPSRC projects on Offshore Robotics for Certification of Assets (ORCA) [EP/R026173/1] and End-to-End Conceptual Guarding of Neural Architectures (EnnCore) [EP/T026995/1], the UK Dstl projects on Test Coverage Metrics for Artificial Intelligence and Safety Argument for Learning-enabled Autonomous Underwater Vehicles (SOLITUDE), and ORCA Partnership Resource Fund (PRF) Towards the Accountable and Explainable Learning-enabled Autonomous Robotic Systems (AELARS). This project has received funding from the European Union’s Horizon 2020 research and innovation programme under grant agreement No 956123.

\balance
\bibliography{references}
\bibliographystyle{plain}

\end{document}